\documentclass{article}

\usepackage{microtype}
\usepackage{graphicx}
\usepackage{subfigure}
\usepackage{booktabs} % for professional tables

\usepackage{hyperref}
\usepackage{ulem}
\usepackage{wrapfig}

\usepackage[accepted]{icml2024}

\usepackage{amsmath}
\usepackage{amssymb}
\usepackage{mathtools}
\usepackage{amsthm}
\usepackage{enumitem}

\usepackage{booktabs}
\usepackage{multirow}
\usepackage{color}

\usepackage[capitalize,noabbrev]{cleveref}

\theoremstyle{plain}

\theoremstyle{definition}

\theoremstyle{remark}

\usepackage[textsize=tiny]{todonotes}
\usepackage{xspace}

\definecolor{commentgreen}{RGB}{2,200,10}
\definecolor{weborange}{RGB}{255,165,0}
\definecolor{darkteal}{RGB}{4,93,93}
\definecolor{commentblue}{rgb}{0.21,0.49,0.74}

\newcommand{\algoName}[0]{\textsc{Flextron}\xspace}

\usepackage{ifthen}

\newcommand{\final}{0}

\ifthenelse{\equal{\final}{1}}{\renewcommand{\todo}[1]{}}{}

\newcommand{\ruisi}   [1]{{{\color{commentgreen}(ruisi) #1}}}
\ifthenelse{\equal{\final}{1}}{\renewcommand{\ruisi}[1]{}}{}

\newcommand{\saurav}   [1]{{{\color{weborange}(saurav) #1}}}

\ifthenelse{\equal{\final}{1}}{\renewcommand{\saurav}[1]{}}{}

\icmltitlerunning{\algoName: Many-in-One Flexible Large Language Model}

\begin{document}

\twocolumn[
\icmltitle{\algoName: Many-in-One Flexible Large Language Model}
% \icmltitle{\algoName: Many in One Flexible Large Language Model}


\icmlsetsymbol{equal}{*}
\begin{icmlauthorlist}
\icmlauthor{Ruisi Cai}{yyy,ut}
\icmlauthor{Saurav Muralidharan}{yyy}
\icmlauthor{Greg Heinrich}{yyy}
\icmlauthor{Hongxu Yin}{yyy}

\icmlauthor{Zhangyang Wang}{ut}
\icmlauthor{Jan Kautz}{yyy}
\icmlauthor{Pavlo Molchanov}{yyy}
\end{icmlauthorlist}

\icmlaffiliation{yyy}{NVIDIA}
\icmlaffiliation{ut}{The University of Texas at Austin}

\icmlcorrespondingauthor{Saurav Muralidharan}{sauravm@nvidia.com}

\icmlkeywords{Machine Learning, ICML}

\vskip 0.3in
]



\printAffiliationsAndNotice{}  % leave blank if no need to mention equal contribution

\begin{abstract}
Training modern LLMs is extremely resource intensive, and customizing them for various deployment scenarios characterized by limited compute and memory resources through repeated training is impractical. In this paper, we introduce \algoName, a network architecture and post-training model optimization framework supporting flexible model deployment. The \algoName architecture utilizes a nested elastic structure to rapidly adapt to specific user-defined latency and accuracy targets during inference with no additional fine-tuning required. It is also input-adaptive, and can automatically route tokens through its sub-networks for improved performance and efficiency. We present a sample-efficient training method and associated routing algorithms for systematically transforming an existing trained LLM into a \algoName model.
We evaluate \algoName on the GPT-3 and LLama-2 family of LLMs, and demonstrate superior performance over multiple end-to-end trained variants and other state-of-the-art elastic networks, all with a single pretraining run that consumes a mere $7.63\%$ tokens compared to original pretraining.

\end{abstract}

\section{Introduction}

Large language models (LLMs) have revolutionized real-world natural language processing applications and have showed impressive proficiency in understanding difficult contexts~\cite{brown2020language, openai2023gpt4, wei2022chain, touvron2023llama}. Nonetheless, the substantial size of these models, typically running into several billion parameters, imposes significant constraints on their utilization in scenarios characterized by limited memory and computational resources. To address this limitation, model providers typically train multiple model variants for users to choose from (depending on system memory and computational constraints) before trying to find model(s) satisfying the trade-off between efficiency and accuracy. For instance, the Llama-2 model family~\cite{touvron2023llama} includes three different variants with 7 billion, 13 billion, and 70 billion parameters, while the Pythia family~\cite{biderman2023pythia} offers a selection of eight models with sizes ranging from 80 million to 12 billion parameters.

\begin{figure}[tb]
    \centering
    \includegraphics[width=1.0\linewidth]{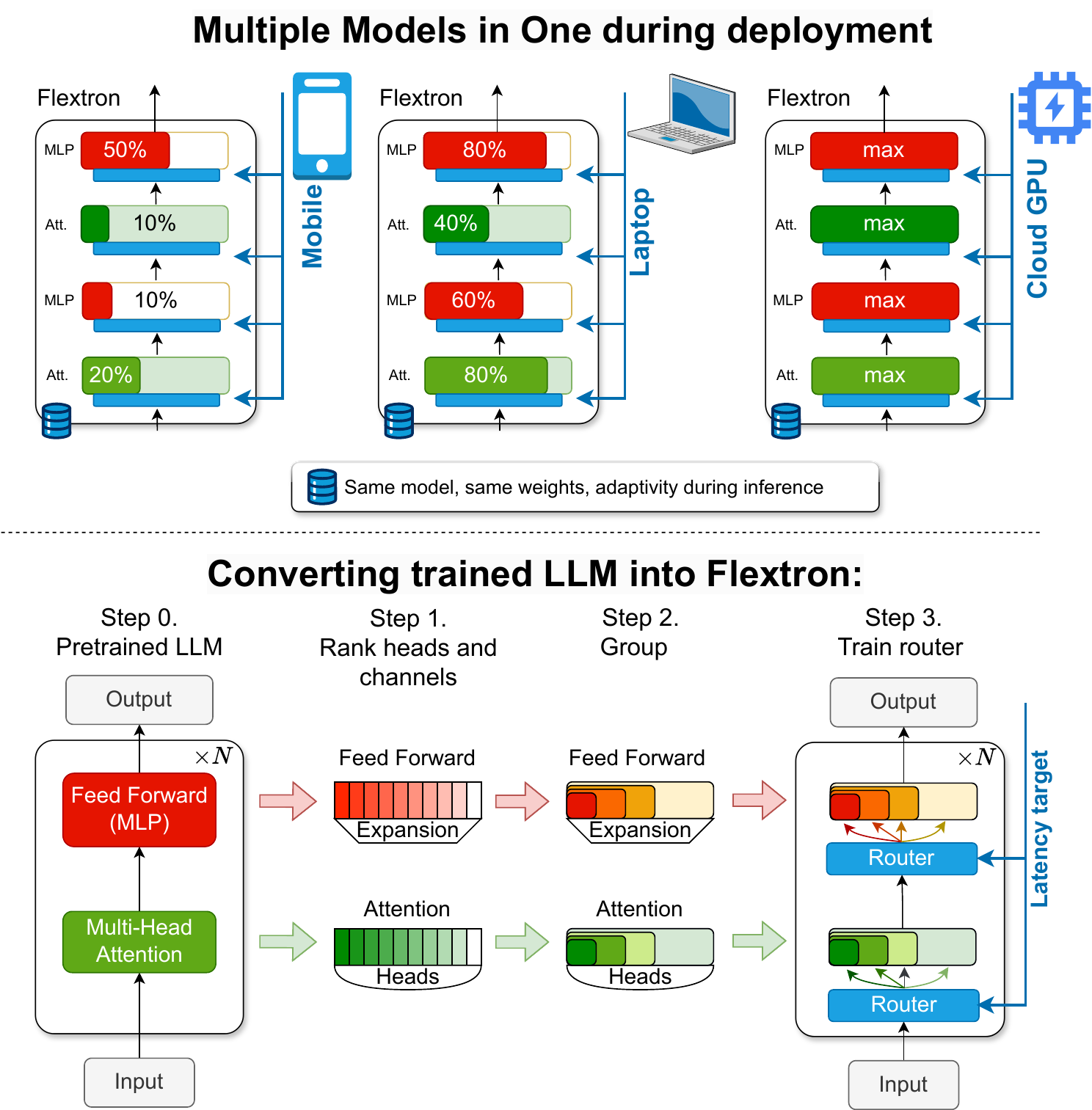}
    \caption{High-level overview of the \algoName framework.
    As shown in the top half of the Figure, \algoName enables fast, zero-shot generation of hardware and input-adaptive sub-networks targeting various accuracy, latency and parameter constraints.
    The bottom half of the figure demonstrates how we convert a trained LLM into an elastic network with input-adaptive routing.
    }
    \label{fig:overview}
\end{figure}

Training multiple multi-billion parameter models is demanding in time, data, and resources. Adopting a single, customizable model with multiple sub-networks for varied budgets, as seen in Once-for-all~\cite{cai2019once},  SortedNet~\cite{valipour2023sortednet}, Matformer~\cite{kudugunta2023matformer}, and \cite{stamoulis2019single}, simplifies this. These models typically use a \textit{supernet} with elastic, nested components, but require non-standard, costly architectures with even longer training than a single model.

Mixture-of-Expert (MoE) networks, while more efficient than dense models~\cite{fedus2022switch,riquelme2021scaling,jiang2024mixtral}, are generally restricted to feedforward layers and fixed budgets. The Pathways architecture~\cite{dean2021introducing,zhou2022mixture} highlights the potential of heterogeneous expert networks. We advocate for input-adaptive sub-network selection of different sizes to maximize performance and efficiency.

In this paper, we present \algoName, a network architecture and a post-training model optimization framework that takes the best from MoEs, elastic models and dynamic inference. The architecture extends the idea of MoE to attention and feed forward layers. Experts are heterogeneous and have different sizes via a nested elastic structure to support efficient model storage, memory bandwidth savings and ease-of-use. Particular experts are selected via a router conditioned on input data and target deployment constraints.
\algoName is a single model that provides \textit{Multiple Models in One} during deployment with no additional finetuning. Finally, we present a framework where a standard trained LLM such as GPT-3 and Llama-2 can be efficiently converted to \algoName while using a small fraction of the training time.  
Figure~\ref{fig:overview} provides a high-level overview. 

We found that training a router that allows adaptive computation is challenging due to gradient vanishing. Similar issues arise in normal MoE training, known as expert collapse~\cite{chi2022representation}, where routers constantly pick the same path or learn similar experts. To address this issue, we propose to train a \textit{Surrogate Model (SM)} that predicts an LLM's language loss value given only router choices; once trained, we freeze it and tune routers to minimize the language loss solely on SM feedback.

This paper makes the following contributions:
\begin{itemize}
    \item A novel architecture, called \algoName, that flexibly adapts to different latency and accuracy targets during inference with no additional fine-tuning. 
    \item A post-training optimization framework for systematically transforming existing trained LLMs into dynamic (input-adaptive) elastic networks. 
    \item New static and dynamic routing algorithms that automatically select the optimal sub-network given a latency target and/or input token. We introduce a novel surrogate model for effective training of our routers. 
    % \vspace{-0.3em}
    \item An efficient sampling-based training method for elastic networks that requires significantly less compute than existing methods.
\end{itemize}

\section{Background and Notation}
\label{sec:background}

Given a model with $N$ layers, each layer can be formalized as $\mathbf{Y}_{i} = f_{i}(\mathbf{X}_{i}, \boldsymbol{W}_{i})$, where $i \in [1, N]$ refers to the layer index, $\mathbf{X}_{i}$ denotes the layer input, with dimensions of $B\times~C$ representing batch $\times$ embedding dimension, and $\boldsymbol{W}_{i}$ denotes the parameters of the layer. We define an \textit{elastic network} as one that can flexibly adapt its layers to target specific user-defined objectives such as latency, memory, accuracy, etc. In this paper, we define each layer of an elastic network as follows: $\mathbf{Y}_{i} = f_{i}(\mathbf{X}_{i}, \boldsymbol{W}^{j}_{i})$ where each $\boldsymbol{W}_i^{j}, j \in [1, K]$ represents a different parameter matrix for the same operation $f_{i}$ for layer $i$. By substituting the original layer with a candidate layer, we are able to generate an exponential number of elastic sub-networks ($K^{N}$ choices for the formulation above, assuming $K$ candidates per layer), each with different runtime and accuracy characteristics.

\paragraph{Elastic Multi-Layer Perceptron (MLP).} \algoName models utilize a nested structure for elastic MLP layers, inspired by the Matformer work~\cite{kudugunta2023matformer}. Nesting enables hidden neurons to be shared between layer-wise candidates using simple indexing operations, saving memory and improving efficiency. Formally, elastic MLP candidates with $2$ linear layers have the following format:
\begin{equation}
    \operatorname{MLP}^j(x) = \sigma\bigg(\mathbf{X} \cdot \big( \mathbf{I}_{d_j} \boldsymbol{W}^{(1)} \big)^T\bigg) \cdot \big( \mathbf{I}_{d_j}\boldsymbol{W}^{(2)} \big),
\end{equation}
where, $\mathbf{I}_{d_j}$ is a diagonal matrix of size $D\times D$  where the first $d_j$ diagonal elements being $1$ and the rest being $0$, with $D$ being the maximum hidden dimension. In this way, the $j^{th}$ MLP candidate will only utilize the first $d_j$ hidden neurons from the corresponding shared matrices $\boldsymbol{W}$. $\boldsymbol{W}^{(1)}$ and $\boldsymbol{W}^{(2)}$ are the associated two weight matrices in MLP layers, with $\boldsymbol{W}^{(1)}, \boldsymbol{W}^{(2)} \in \mathbb{R}^{D\times C}$; $\sigma(\cdot)$ refers to the non-linear activation function. 
For implementation, the diagonal matrix $\mathbf{I}$ can be replaced with a slicing operator that selects only the first $d_j$ rows: $\mathbf{I}_{d_j} \boldsymbol{W}^{(1)} = \boldsymbol{W}^{(1)}[0:d_j,:]$. We constrain $d_1 < d_2 < ... < d_K$, where $d_K=D$, to formulate the nested structure of $K$ experts. Note that the MLP can take a more complex form when employing SwiGLU activation.

\paragraph{Elastic Multi-Head Attention (MHA).}
MHA layers constitute a significant proportion of LLM runtime and memory usage (for KV cache), and making them elastic will improve overall efficiency. To the best of our knowledge, \algoName is the first work that supports both elastic MLP and elastic MHA layers, enabling a richer candidate operation search space. An elastic MHA candidate uses a subset of attention heads.
Formally, given hidden states $\mathbf{X}$, we define elastic MHA as follows:
\begin{equation}
\begin{aligned}
    \operatorname{MHA}^j(x) &= \operatorname{Concat}(\text{head}_1, ... \text{head}_{d_j}) \cdot \big(\mathbf{I}_{d_jH} \boldsymbol{W}^O \big), \\
    &\text{head}_i = \operatorname{Attn}(\mathbf{X} \boldsymbol{W}^{Q,i}, \mathbf{X} \boldsymbol{W}^{K,i}, \mathbf{X} \boldsymbol{W}^{V,i}), \\
\end{aligned}
\end{equation}
where, $\mathbf{I}_{d_jH}$ is a diagonal matrix with the first $d_jH$ elements being 1, and the rest are 0s; $d_j$ - number of heads selected, $H$ - size of a single head, $L$ - total number of heads ; $\boldsymbol{W}^{Q,i}, \boldsymbol{W}^{K,i}, \boldsymbol{W}^{V,i} \in \mathbb{R}^{H\times C}$ and $\boldsymbol{W}^O \in \mathbb{R}^{LH\times C}$. Different heads can be computed/selected via weight slicing.

\section{\algoName Framework}

We now describe the elastic network continued-training (CT) process, and provide more details on automatic sub-network selection from the trained elastic network.

\subsection{Elastic Network Continued-Training}
\label{sec:method:training}

We start the elastic continued-training process by taking an existing trained LLM and performing importance ranking for each neuron/head. Here, using a small set of data samples, we compute the importance of each neuron/head based on the accumulated magnitude of activations. For MHA layers, the importance of each head is calculated as 
\begin{equation}
\label{eq:permute_head}
F_{\text{head}}^{(i)} = \sum_{\mathbf{X}} \Vert \operatorname{Attn}(\mathbf{X}\boldsymbol{W}^{Q,i}, \mathbf{X}\boldsymbol{W}^{K,i}, \mathbf{X}\boldsymbol{W}^{V,i}) \Vert_1.
\end{equation}
For MLP layers:  
\begin{equation}
\label{eq:permute_mlp}
F_{\text{neuron}}^{(i)} = \sum_{\mathbf{X}} \Vert \mathbf{X} \big(\boldsymbol{W}^{(1), r}\big)^T \Vert_1,
\end{equation}
here $\boldsymbol{W}^{(1),{r}}$ refers to the $r^\text{th}$ row of the weight matrix $\boldsymbol{W^{(1)}}$. In practice, only a small dataset comprising $512$ samples is sufficient (see Section~\ref{sec:results} for more details).
Once importance is computed, we permute the respective weight matrices in the MLP and MHA layers such that neurons/heads are stored in decreasing order of importance for every individual layer. Sub-networks can now be constructed by simply indexing the first several neurons/heads in each layer, thus preserving essential knowledge encoded in important channels. In this way, we construct nested elastic layers with parameter sharing, with channels/heads sorted by importance, such that the first channels are the most important.  

\begin{figure}[tb]
    \centering
    \vspace{-2mm}
    \includegraphics[width=0.75\linewidth]{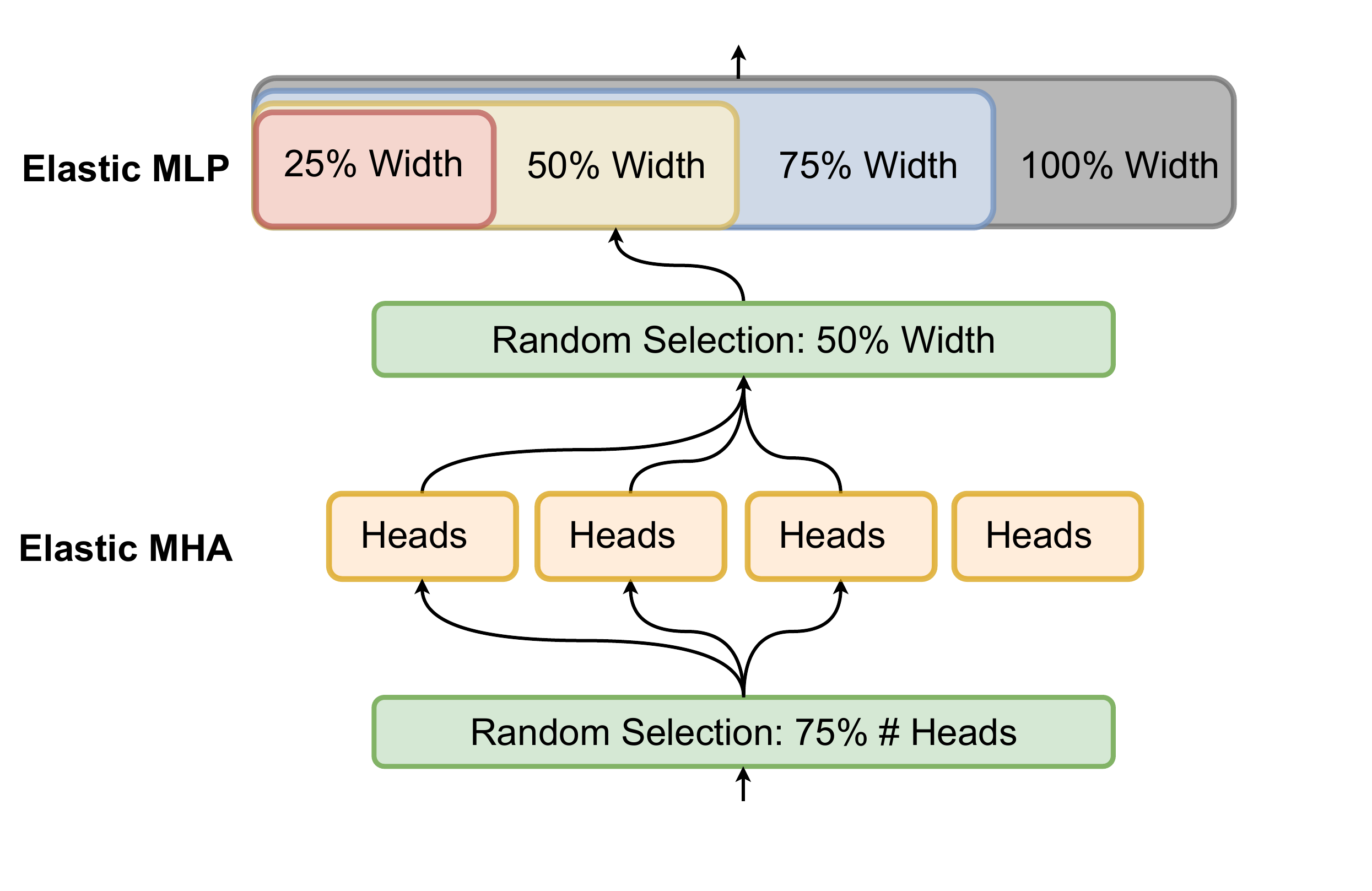}
    \vspace{-5mm}
    \caption{Illustration of the elastic continued-training phase.}
    \label{fig:pretraining}
\end{figure}

Next, we train all elastic network candidates simultaneously using a combined loss term as in~\cite{kudugunta2023matformer}. Since the number of such candidates can be prohibitively large (for example, there are $4^{64}$ possible combinations for the 32-layer LLaMa2-7B model~\cite{touvron2023llama}), we randomly sample a smaller subset of $k$ networks from the candidate pool to keep the total pretraining time tractable. Specifically, we randomly generate a one-hot vector $s_i$ for each layer $i$ and use it to construct a candidate network $\mathcal{M}_j$; here, $s_i \in \mathbb{R}^{K_i}$ and $K_i$ represents the number of candidate MLP/MHA operations in layer $i$. $\mathcal{M}_j$ is the random model indexed by $j$, where $j \in [0, K-1]$.
The training loss is:
\begin{equation}
    \mathcal{L}_\text{joint} = \sum^{k-1}_{j=0} \mathcal{L}(\mathcal{M}_j(\mathbf{x}),\mathbf{y}),
\end{equation}
Figure~\ref{fig:pretraining} provides an overview of elastic continued-training with random sampling. We provide additional details on pretraining, including choice of hyper-parameter values and datasets, in Section~\ref{sec:results}.

\begin{figure*}[tb]
    \centering
    \includegraphics[width=1.0\linewidth]{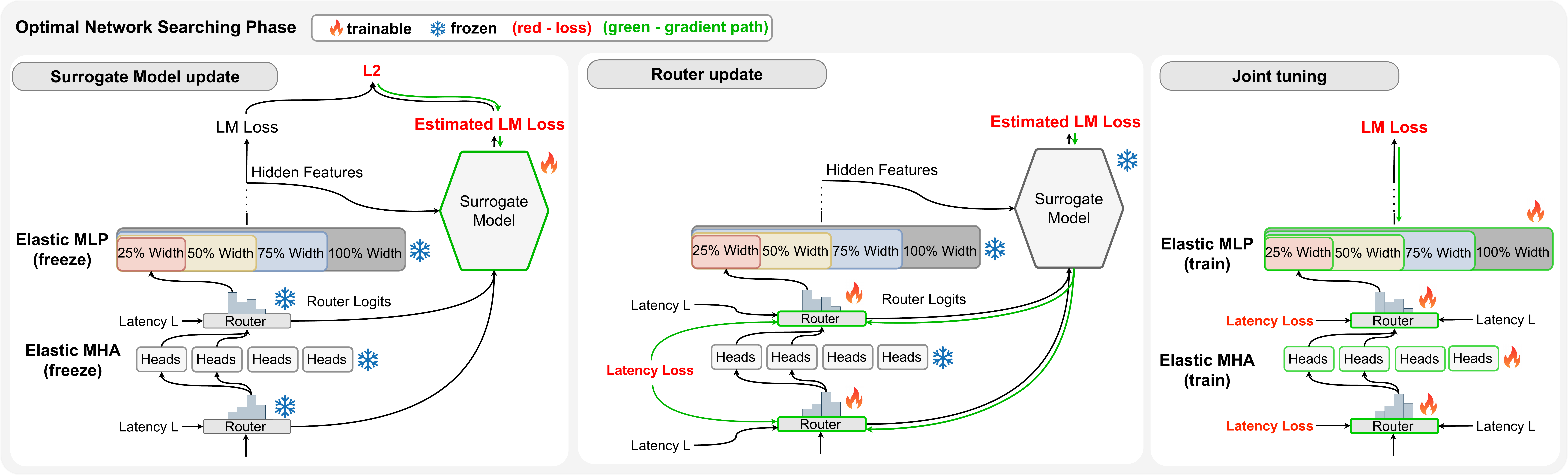}   
    \vspace{-2em}
    \caption{Illustration of how routers are trained via a Surrogate Model (SM). The Surrogate Model is trained to approximate the LLM language loss value given only routers logits. If the error of the SM is smaller than a predefined threshold, the routers are updated. Updates are based on (i) the latency loss, ensuring the requested latency matches the real overall latency via a Lookup Table (LUT), and (ii) the loss from minimization of the SM output. The SM serves as a proxy for the full model's language loss and allows for simpler backpropagation due to its smaller size. Once the routers are trained, we discard the SM and finetune the LLM and routers jointly.}
    \label{fig:train_router}
\end{figure*}

\subsection{Automatic Network Selection}
\label{sec:method:static}
Given a large number of possible sub-networks to choose from, each with different latency, parameter, and accuracy trade-offs, a natural question arises: \textit{can we automatically determine Pareto-optimal sub-networks given specific constraints?}
In this section, we introduce \algoName's router architecture and describe how it helps us automatically select optimal sub-networks for a given constraint. %Figure~\ref{fig:train_router} provides an overview of the router design.

The problem can be formalized as follows:
\begin{equation}
   \begin{aligned}
       \min_{\boldsymbol{S}_t} \  \sum_t  \mathcal{L}_{CE}(\mathcal{M}_{s_t}), & 
       \quad \text{s.t.} \operatorname{Latency}(\mathcal{M}_{s_t}) \leq T_t, \\
       &\quad \mathcal{M}_{s_t} = \mathcal{G}(\mathcal{M}, \boldsymbol{S}_{T_t}),
   \end{aligned}
\end{equation}
where $\mathcal{M}$ is original network topology, $T_t$ refers to a latency constraint of index $t$, and $\boldsymbol{S}_{T_t}$ denotes the related selection matrix; $\mathcal{M}_{s_t}$ defines the selected topology based on latency constraint, $\mathcal{G}(\cdot)$ is a function for selecting network topology; $\mathcal{L}_{CE}$ refers to the cross-entropy loss. We use a Lagrange multiplier and impose a constraint to convert the aforementioned optimization problem into directly minimizing the following loss term:
\begin{equation}
   \mathcal{L} = \sum_t \mathcal{L}_{CE}(\mathcal{M}_{s_t}) + \lambda \cdot \mathcal{T}_{T}(\mathcal{M}_{s_t}),
\end{equation}
where $\mathcal{T}_{T}$ represents the target constraint loss. In what follows as an example, we explain \textit{latency loss} between the constraint $T_t$ and actual model latency $\operatorname{Latency}(\mathcal{M}_{s_t})$:
\begin{equation}
   \mathcal{T}_{T}(\mathcal{M}_{s_t}) = \sum_t \max (\operatorname{Latency}(\mathcal{M}_{s_t}) - T_t, 0)
\label{eq:latency}
\end{equation}

Note that the loss can also be extended to other constraints such as GPU memory as we show later in our experiments.

The model requires an architecture selection mechanism to support multiple budgets with maximized accuracy. Inspired by MoEs, we use routers. We define two routing scenarios: \textit{static}, where the output depends only on the input latency; and \textit{dynamic}, where it is additionally conditioned on the hidden state.
We observe that training routers, even after the elastic continued-training stage, is challenging due to limited gradient propagation from the final model's output loss. As a remedy, we propose using a surrogate model to predict the LLM's performance based solely on router outputs. Given this prediction, routers can be trained to minimize the expected LM loss of the surrogate model. We provide additional details on the surrogate model in the following sections.

\paragraph{\algoName-Static: Static Model Selection.}
We first tackle the problem of static model selection, which refers to automatically selecting sub-networks given only a target latency $T$ (no input-adaptivity). Here, we insert layer-wise learnable routers; each router takes the latency requirement $T$ as input and outputs the choice $h_i$ for layer $i$, thereby deciding the number of channels/heads to be used for that layer via expert groups.
The router picks the expert with the following formulation:
\begin{align}
    s_i = \operatorname{argmax}(\mathcal{R}_i(T)),
\label{eq:gate}
\end{align}
where $\mathcal{R}$ is a small MLP that embeds a scalar value $T$ (latency) into logits of the size of the predefined set of expert candidates (in our paper selected to be 4).

To provide a strong and stable signal to the router, we propose to use a \textit{Surrogate Model (SM)}. Its task is to predict the value of the full LLM language loss given only logits at the outputs of the routers. It becomes a proxy for the full model output error. Once it is learned, we can optimize routers to minimize the output of the SM. The basic idea is to use the SM as a loss term that can be minimized. The SM is defined as a two layer MLP:
\begin{equation}
\label{eq:sm}
\begin{aligned}
    r =& \operatorname{Concat}(\mathcal{R}_0(T), \mathcal{R}_1(T), ... \mathcal{R}_{N-1}(T)), \\
    &\mathcal{S}(r) = \sigma (r \boldsymbol{W}_{\mathcal{S}_1}^T) \boldsymbol{W}_{\mathcal{S}_2},
\end{aligned}
\end{equation}
where $\boldsymbol{W}_{\mathcal{S}_1}$ and $\boldsymbol{W}_{\mathcal{S}_2}$ are weights of size $\boldsymbol{W}_{\mathcal{S}_1} \in \mathbb{R}^{P \times K\cdot N}$ and $\boldsymbol{W}_{\mathcal{S}_2} \in \mathbb{R}^{P \times 1}$; $P$ is an internal hidden dimension.

\paragraph{\algoName-Adaptive: Dynamic Model Selection.}
Recent work on sparsely-activated MoE networks has demonstrated that an ensemble of different sub-networks (``experts''), each specializing in particular input domains, performs better and more efficiently than dense baselines~\cite{fedus2022switch,zhou2022mixture}.
Drawing inspiration from previous studies, \algoName introduces an input-adaptive routing mechanism to dynamically select optimal sub-networks based on latency and input, reducing memory and communication overheads through weight sharing and array-based indexing.

For input adaptivity, we modify the router design to also incorporate the current hidden states $h_i$ as follows:
\begin{equation}
\label{eq:input-adaptive}
    \begin{aligned}
        s_i =& \operatorname{argmax}(\mathcal{R}_i(T, h_i)), \\
    \text{where} & \  \mathcal{R}_i(T, h_i) = \sigma(T\cdot \boldsymbol{W} +    h_i\boldsymbol{W}_{H_i}^T )\boldsymbol{W}_{\mathcal{R}_i}
    \end{aligned}
\end{equation}
Here, the current hidden features $h_i$ are projected into the embedding space of dimension $U$ by an MLP layer parameterized by $\boldsymbol{W}{H_i}$. Similarly, $T$ is projected via simple scaling of the matrix $\boldsymbol{W}$. We limit $U$ to 128. Token-wise routing decisions are generated by aggregating the latency embedding vector with the hidden feature embedding vector, and passing them through a linear layer. 

We also extend the surrogate model format described in Eq. (\ref{eq:sm}) to additionally incorporate the final hidden states $h_N$. Hidden states are projected to the dimension of $P$ using a linear matrix. This projection is then summed with the latency embedding before applying the activation function.

\noindent \textbf{Training.} Figure~\ref{fig:train_router} provides an overview of training routers via SM. Initially, the main LLM is frozen. Routers are always updated with gradients from the latency loss defined in Eq. (\ref{eq:latency}). The Surrogate Model is updated to minimize the predicted LM loss (via the MSE objective). If the MSE is below a predefined threshold, then routers are additionally updated with gradients from the output of the SM to minimize the predicted LM loss. In this way, routers learn to minimize the LM loss in an indirect way, via SM. Once the routers are trained, we disregard the SM and fine-tune both the routers and the LLM parameters.

\section{Experiments}
\label{sec:results}

\subsection{Experimental Settings}

\paragraph{Model and Dataset.} We perform our evaluation on the GPT3 and Llama2~\cite{touvron2023llama} model families. GPT3 is a representative multilingual model family~\cite{shoeybi2019megatron} with $2$ and $8$ billion parameter variants (among others); these are pretrained with the NeMo framework~\cite{kuchaiev2019nemo}. The total number of trainable parameters for GPT3-$2$B is $2$ billion, with $1.2$ billion non-embedding parameters. The model contains $24$ layers, with a hidden dimension of $2048$. Each MHA layer possesses $16$ heads. 
GPT3-$8$B comprises $8$ billion parameters, of which $6.4$ billion are non-embedding parameters. The model contains $32$ layers, each with a hidden dimension of $4096$. Each MHA layer possesses $32$ heads.
Both models support a maximum context length of $4096$.
GPT3 2B and 8B are trained on $1.1$ trillion tokens, where data is obtained from publicly available data sources, comprising $53$ languages and code~\cite{shoeybi2019megatron}.
We further validate our approach using the Llama2-7B model~\cite{touvron2023llama}, a widely used open-source pre-trained model with $6.5$ billion non-embedding parameters. This model employs a $32$-layer transformer architecture with a hidden dimension of $4096$, incorporating $32$ attention heads for each Multi-Head Attention (MHA) mechanism.
Llama2-7B is trained on $2$ trillion tokens~\cite{touvron2023llama}.
We perform neuron/head sorting, elastic continued-training and router tuning with the same domain data.
For both GPT3 and Llama2, we use $89.9$B tokens for elastic continued training, and $1.049$B tokens for router tuning.
For Llama2, we additionally sample a subset comprising of $2.62$B tokens from the Nemotron-4 curated continued training dataset~\cite{parmar2024nemotron} for joint tuning.

\paragraph{Baselines.} We compare our method with Matformer~\cite{kudugunta2023matformer}, which adopts a nested structure on MLPs, obtaining $4$ variants per MLP layer by training once. To ensure fair comparison, instead of training the Matformer models from scratch, we adopt the pretraining strategy described in Section~\ref{sec:method:training}. We also compare our method with smaller pretrained models trained on the same data and with the same training recipe. We choose GPT$3$-$2$B and GPT$3$-$8$B, our base model, and GPT$3$-$843$M, a smaller version of GPT$3$ with embedding size of $1024$, $24$ layers and $16$ heads.
We additionally compare our method with representative open-source model families, including Pythia~\cite{biderman2023pythia}, OpenLLaMA~\cite{openlm2023openllama}, and models generated by post-hoc compression methods, including Sheared-LLaMA~\cite{xia2023sheared}, Compresso~\cite{guo2023compresso}, LLM-Pruner~\cite{ma2023llm}, SliceGPT~\cite{ashkboos2024slicegpt}, and LaCo~\cite{yang2024laco}.

\paragraph{Training.}
As described in Section~\ref{sec:method:training}, during elastic network pretraining, we first perform importance sorting of each head/neuron in MHA/MLP layers using a tiny fraction ($512$ samples) of the full training set . 
%The samples are randomly collected from the same domain as the pretrained data.
We then perform training of the sorted and permuted elastic model. We use a batch-size of $256$, and tune the model for $80000$ steps. At each step, we randomly construct $3$ sub-models together with the full model; perform gradient accumulation for all $4$ models for a single update.
We perform lightweight tuning for automatic network selection: we freeze the backbone parameters and only tune the routers and surrogate models for $1000$ steps using a batch size of $256$. For static router tuning, we observe a consistent performance ranking over multiple data domains for sub-models, and thus use only single domain data (Wikipedia~\cite{wikidump}). During the input-adaptive router training, which is harder, we use the subset of pretraining dataset.

\subsection{Results}

\paragraph{\algoName Performance.}

\begin{table*}[htb]
\caption{Downstream task evaluation of \algoName family models and comparison with representative open-source models and compression methods. We report the zero-shot performance of ARC-easy~\cite{clark2018think}, LAMBADA~\cite{paperno-EtAl:2016:P16-1}, PIQA~\cite{Bisk2020}, and WinoGrande~\cite{sakaguchi2021winogrande}. We also report the 5-shot performance of MMLU\cite{hendrycks2020measuring}, and the 10-shot performance of HellaSwag~\cite{zellers2019hellaswag}. Here, \texttt{\#params} refers to the number of \textit{non-embedding} parameters. Note that for dynamic \algoName models, we use the averaged number of activated non-embedding parameters. 
\label{table:downstream}}
\resizebox{\textwidth}{!}{
\begin{tabular}{@{}lllccccccc@{}}
\toprule
 &  & \# Params & ARC-E & LAMBADA &  PIQA & Winogrande & MMLU (5) & Hellaswag (10) & Avg.\\ \midrule
 & Full & 6.4 B & 71.7\% & 69.7\% & 79.4\% &	68.8\% & 35.4\% & 75.9\% & 66.8\%\\ \cmidrule{2-10}
 & Static-0.7$\times$ &  4.1 B & 66.7\% & 62.9\% & 75.1\% & 63.9\% & 28.7\% & 70.6\% & 61.3\%\\
 & Dynamic-0.7$\times$  & 4.3 B  & 67.0\% & 64.8\% & 75.9\% & 64.1\% & 30.0\% & 70.4\% & 62.0\%\\ \cmidrule{2-10}
 & Static-0.6$\times$  & 3.9 B  &  66.2\% & 62.8\% & 75.6\% & 62.7\% & 28.8\% & 68.8\% & 60.8\%\\ 
 & Dynamic-0.6$\times$  & 3.9 B &   66.2\% & 63.7\% & 76.1\% & 62.7\% & 29.1\% & 69.2\% & 61.2\%\\ \cmidrule{2-10}
 & Static-0.5$\times$  & 3.4 B &  64.2\% & 62.0\% & 74.9\% & 61.7\% & 25.1\% & 66.8\% & 59.1\%\\
\multirow{-8}{*}{\algoName-8B} & Dynamic-0.5$\times$ & 3.3 B & 65.0\% & 62.5\% & 75.8\% & 61.8\% & 27.1\% & 67.8\% & 60.0\%\\
\midrule
 & Full & 6.5 B  &  75.1\%	& 71.5\% & 77.5\%	& 69.1\% & 45.1\% & 78.1\% & 69.4\% \\ \cmidrule{2-10}
 & Static-0.7$\times$ & 4.2 B & 65.8\% & 64.2\% & 75.6\% & 62.3\% & 41.9\% & 67.1\% & 62.8\%\\
 & Dynamic-0.7$\times$  & 4.1 B & 68.6\%  & 65.1\% & 76.1\% & 63.7\% & 42.2\% & 69.4\% & 64.2\%\\
 \cmidrule{2-10}
 & Static-0.6$\times$ & 4.0 B & 66.1\% & 63.8\% & 75.0\% & 62.1\% & 37.7\% & 68.0\% & 62.1\%\\
 & Dynamic-0.6$\times$  & 3.9 B  & 67.1\% & 63.8\% & 74.9\% & 62.2\% & 39.4\% & 69.7\% & 62.8\%\\
 \cmidrule{2-10}
 & Static-0.5$\times$ & 3.5 B & 65.9\% & 61.7\% & 74.8\% & 61.9\% & 35.9\% & 67.6\% & 61.3\%\\
\multirow{-8}{*}{\algoName-Llama2-7B} & Dynamic-0.5$\times$  & 3.4 B & 66.5\% & 62.9\% & 74.1\% & 62.0\% & 36.8\% & 68.5\% & 61.8\% \\
\midrule
 & Llama2-7B & 6.5 B & 75.2\%  & 68.2\% & 78.8\% & 69.2\% & 45.3\% & 78.6\% & 69.2\% \\ 
 & OpenLLaMA-3Bv2 & 3.2 B & 63.7\% & 59.1\% & 78.1\% & 63.3\% &25.7\% & 71.6\% & 60.3\%\\
 & OpenLLaMA-7Bv2 & 6.5 B & 69.5\% & 63.8\% & 79.9\% & 66.0\% & 40.4\% & 76.6\% & 66.0\% \\
 & GPT3-8B & 6.4 B & 70.1\% & 70.5\% & 79.7\% & 69.8\%& 40.2\% & 77.7\% & 68.0\% \\
 & Pythia-1.4B & 1.2 B & 53.9\% & 46.8\% & 70.6\% & 57.1\% & 25.6\% & 52.2\% & 51.0\%\\
 & Pythia-2.8B & 2.5 B & 57.9\% & 50.1\% & 73.8\% & 58.6\% & 26.8\% & 60.0\% & 54.5\%\\ 
\multirow{-7}{*}{Open-Source} & Pythia-6.9B & 6.4 B  & 60.2\% & 47.1\% & 75.2\% & 59.9\% & 25.5\% & 64.4\% & 55.4\%\\ 
\midrule
 & Sheared-LLaMA-1.3B & 1.2 B & 61.5\% & 61.0\% & 73.4\% & 57.9\% & 25.7\% & 60.7\% & 56.7\% \\
\multirow{5}{*}{Compressed} & Sheared-LLaMA-2.7B & 2.5 B & 67.0\% & 68.4\% & 75.8\% & 64.2\% & 26.4\% & 70.8\% & 62.1\%
\\ 
& NutePrune & 3.2 B & 51.7\% & - & 71.0\% & 57.5\% & - & 55.9\% & -\\
& LLM-Pruner & 4.5 B & 59.2\% & - & 73.4\% & 64.2\% & 23.9\% & 56.5\% & - \\
& Compresso & 4.5 B & 66.0\% &  - & 72.9\% & 63.4\% & 25.9\% & - & - \\
& LaCo & 4.7 B & - & - & 69.8\% & - & 26.5\% & 55.7\% & -  \\
& SliceGPT & 4.8 B & - & - & 66.2\% & - & 28.9\% & 50.3\% & - \\
\bottomrule
\end{tabular}}
\end{table*}

We validate the effectiveness of \algoName on multiple downstream tasks in Table~\ref{table:downstream}. These tasks include:  ARC-easy~\cite{clark2018think}, LAMBADA~\cite{paperno-EtAl:2016:P16-1}, PIQA~\cite{Bisk2020}, WinoGrande~\cite{sakaguchi2021winogrande}, MMLU\cite{hendrycks2020measuring}, and HellaSwag~\cite{zellers2019hellaswag}. We follow the common choice in~\citet{xia2023sheared} and report the 5-shot and 10-shot performance for MMLU and Hellaswag, respectively. We report zero-shot performance for other tasks. 
In Table~\ref{table:downstream}, \algoName-8B and \algoName-Llama2-7B denote the Flextron models built upon GPT3-8B and Llama2-7B, respectively. ``Dynamic'' refers to the model with input-adaptive router while ``static'' represent the static case where all tokens select the same sub-network given the latency/memory requirements. $\times$ suffix indicates the remaining latency of the model. 
\begin{figure}[htb]
    \includegraphics[width=\linewidth]{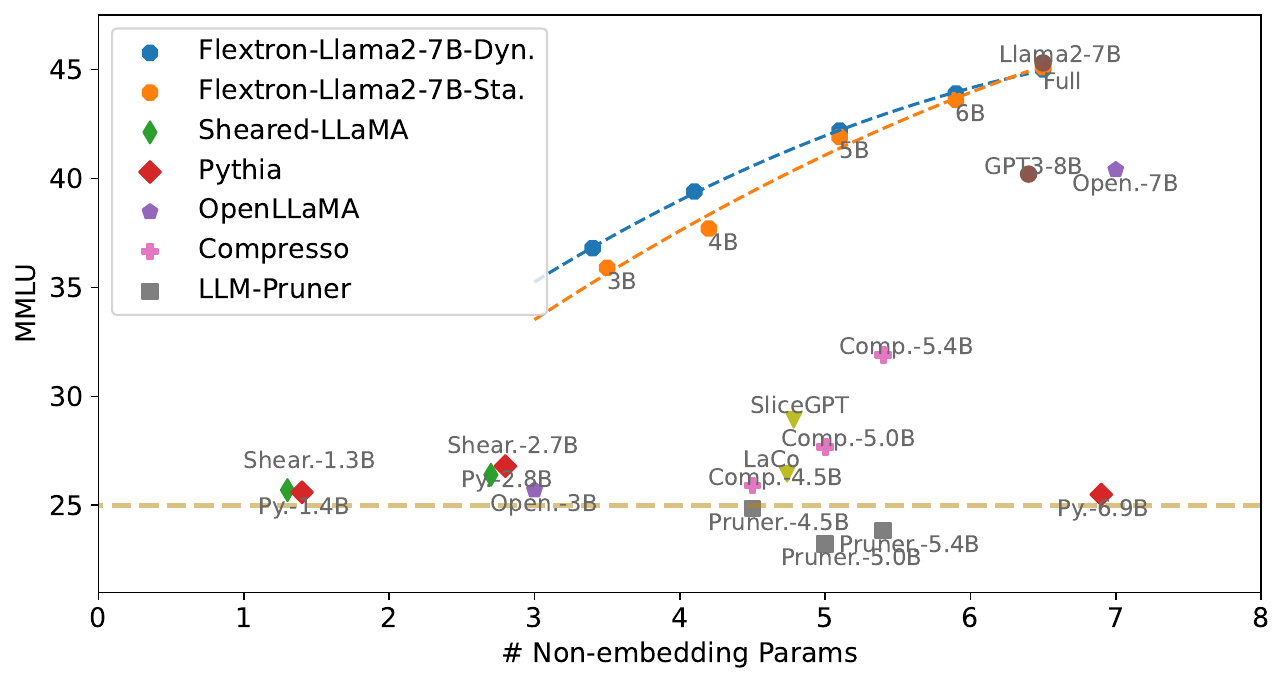}
    \vspace{-1.5em}
    \caption{The Flextron-Llama2-7B model family demonstrates superior MMLU~\cite{hendrycks2020measuring} performance compared to both open-source models and existing post-hoc compression methods. Specifically, we compare against models from the Pythia~\cite{biderman2023pythia} family and the OpenLLaMA-v2~\cite{openlm2023openllama} family. Additionally, our method is compared with Sheared-LLaMA~\cite{xia2023sheared}, Compresso~\cite{guo2023compresso}, LLM-Pruner~\cite{ma2023llm}, SliceGPT~\cite{ashkboos2024slicegpt}, and LaCo~\cite{yang2024laco}. $\times$ suffix indicates the remaining latency of the model.}
    \vspace{-0.5em}
\end{figure}

\begin{figure*}[htb]
    \centering
    \includegraphics[width=0.8\linewidth]{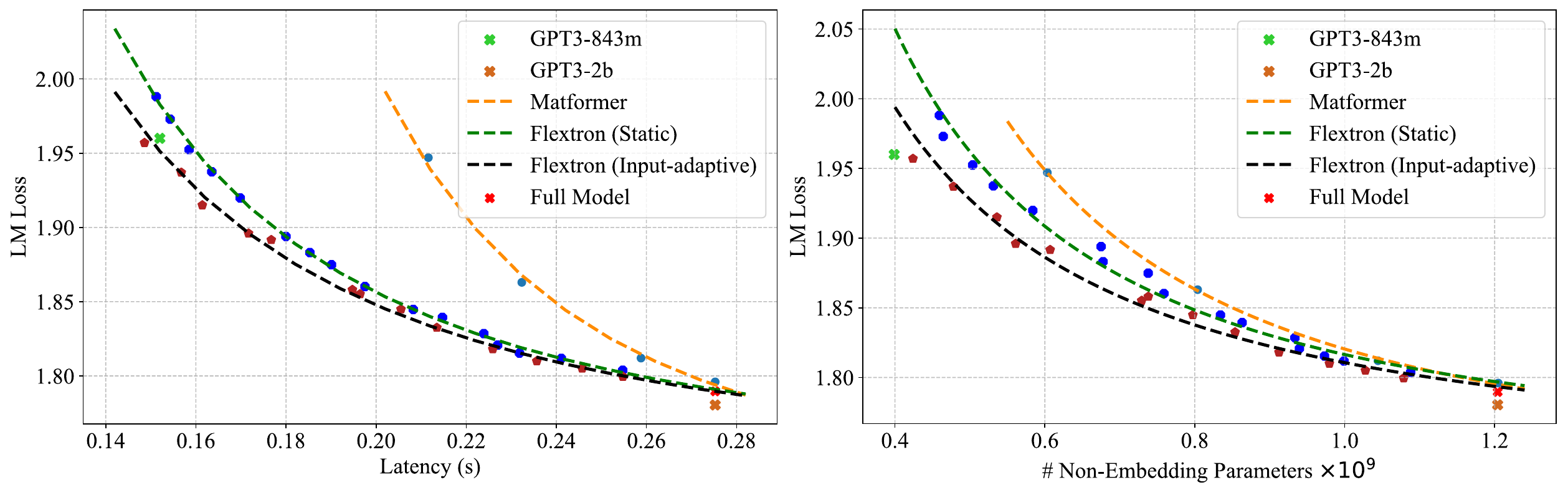}
    \vspace{-0.5em}
    \caption{Pareto curves for language modeling loss vs latency (left) and \# non-embedding parameters (right). The curve is fitted by the model scaling equation. \algoName achieves superior performance to Matformer and even end-to-end-trained smaller models (843M). The performance of the model is evaluated by language modeling validation loss and averaged over $7$ representative datasets: ($1$) English datasets:Arxiv, Books3~\cite{pile}, Wikipedia~\cite{wikidump}, ($2$) multilingual datasets: Korean, German languages, and ($3$) code data: HTML, JAVA. We measure model latency with the Megatron framework~\cite{shoeybi2019megatron} using a batch size of 2 and sequence length of 4096 in the context prefilling stage on NVIDIA A100 GPU.}
    \label{fig:pareto}
    \vspace{-1em}
\end{figure*}

Furthermore, we measure the latency of the \algoName models in Table~\ref{table:latency}, with latency measured using TensorRT-LLM~\cite{tensorrtlm}. It is worth noting that \algoName-8B models are multilingual models with a vocabulary size of $320,000$, and the embedding operation incurs a latency of $1.82$s, constituting $17.4\%$ of full latency. For comparison, the embedding layer of Llama2-7B incurs a latency of $0.69$s, constituting $7.2\%$ of the full latency. All results are tested on the NVIDIA A100 80GB GPU, with latency measured when the prompting length and generation length is set to 8 and 512, respectively. We use a batch size of 1. 

\begin{table}[htb]
\centering
\caption{Latency of \algoName family  models. The latency is measured based on TensorRT-LLM~\cite{tensorrtlm} and NVIDIA A100 80GB GPU. We measure the latency when the prompting length and generation length is set to 8 and 512, respectively. We use the batch size of 1. The reported numbers present \textit{(\# non-embedding params) / (latency)}.\label{table:latency}}
\resizebox{\columnwidth}{!}{
\begin{tabular}{@{}lcccc@{}}
\toprule
& Full & 0.7$\times$ & 0.6$\times$ & 0.5$\times$ \\ 
\midrule
\algoName-8B & 6.4B / 10.43s & 4.1B / 8.02s & 3.9B / 6.39s & 3.4B / 5.48s \\
\algoName-Llama2-7B & 6.5B /\ \ \  9.64s & 4.1B / 7.09s & 3.9B / 5.41s & 3.4B / 4.91s\\ \bottomrule
\end{tabular}}
\end{table}

\paragraph{Neural Scaling Laws.} 

Recent work~\cite{kaplan2020scaling,hoffmann2022training} has empirically demonstrated scaling laws for LLMs with respect to model size. Specifically, model capacity scales as follows: 
\begin{equation}\label{eqa:scaling_law}
    L(N) = (N/N_c)^{-\alpha_N} + E_N,
\end{equation}
here, $N$ denotes the number of non-embedding model parameters, and $N_c$, $\alpha_N$, and $E_N$ are model-dependent coefficients. This curve typically uses multiple \textit{independently} trained models to capture the correlation between model size and validation loss. For \algoName, we extend the model scaling law along two dimensions: ($1$) we observe that the model's capacity, which grows with the number of sub-model parameters, follows the existing model scaling law, and ($2$) we establish a power law relationship between the model's capacity and the input latency.

Figure~\ref{fig:pareto} plots the trade-off between validation loss and latency (left) / number of non-embedding parameters (right) for both \algoName-Static and input-adaptive \algoName-Adaptive routing of the trained elastic model. 
MHA layers typically introduce fewer parameters but incur high latency; as such, elastic MHA is favorable in the higher latency regimes. This is evident when comparing \algoName's performance to Matformer~\cite{kudugunta2023matformer}, which only leverages elastic MLP.
In Figure~\ref{fig:pareto}, we fit the data points of sub-networks with Equation~\ref{eqa:scaling_law}, and provide the fitted parameters for the scaling equation in Table~\ref{table:fitting}; this suggests a useful guideline for model practitioners to choose the proper model that simultaneously meets latency, number of parameters, and model capacity constraints. 
\vspace{-1em}
\begin{table}[htb]
\centering
\vspace{-0.5em}
\caption{Fitted parameters for Equation~\ref{eqa:scaling_law}. \label{table:fitting}}
    \resizebox{.8\columnwidth}{!}{
\begin{tabular}{@{}llll@{}}
\toprule
 & $N_C$ & $\alpha_N$ & $E_N$ \\ 
\midrule
Matformer~\cite{kudugunta2023matformer} & $1.680$ & $52.74$ & $1.729$ \\
\algoName (Static) & $1.465$ & $38.57$ & $1.733$ \\
\algoName (Input-adaptive) & $1.289$ & $25.52$ & $1.729$ \\
\bottomrule
\vspace{-2.5em}
\end{tabular}}
\end{table}

\paragraph{Training Efficiency.}
\algoName demonstrates excellent training efficiency, as detailed in Table~\ref{table:training_efficiency}. 
During elastic continued-training, we utilize only $89.9$ billion tokens for both the GPT3 and LLama2 models, while for router tuning, we use $1.049$ billion tokens.

\begin{table}[htb]
\centering
\vspace{-1em}
\caption{Flextron training costs compared to pretraining cost. We report the number of tokens for elastic CT, router tuning and joint tuning (in case of Llama2-7B) to illustrate the training cost.
 \label{table:training_efficiency}}
\resizebox{\columnwidth}{!}{
\begin{tabular}{@{}lccc@{}}
\toprule
 & \multicolumn{2}{c}{Flextron Training Cost} & \multicolumn{1}{c}{\multirow{2}{*}{Pretraining Cost}} \\ \cmidrule(r){2-3}
 & \multicolumn{1}{c}{Elastic Continued-Training} & \multicolumn{1}{c}{Router Tuning} & \multicolumn{1}{c}{} \\ \cmidrule(l){1-4} 
\multicolumn{1}{l}{GPT3} & 89.9 B ($7.54\%$) & 1.049 B ($0.09\%$) & 1.1T \\
\multicolumn{1}{l}{Llama2} & 89.9 B ($4.50\%$) & 1.049 B + 2.62 B ($0.18\%$) & 2T \\ \bottomrule
\vspace{-1.5em}
\end{tabular}}
\end{table}

\section{Analysis}
\subsection{\algoName Learnings \& Insights}\label{sec:insights}

\paragraph{Routers Assign More Computation to Deeper MLP layers.}

\begin{figure}[htb]
    \centering
    \includegraphics[width=1.0\linewidth]{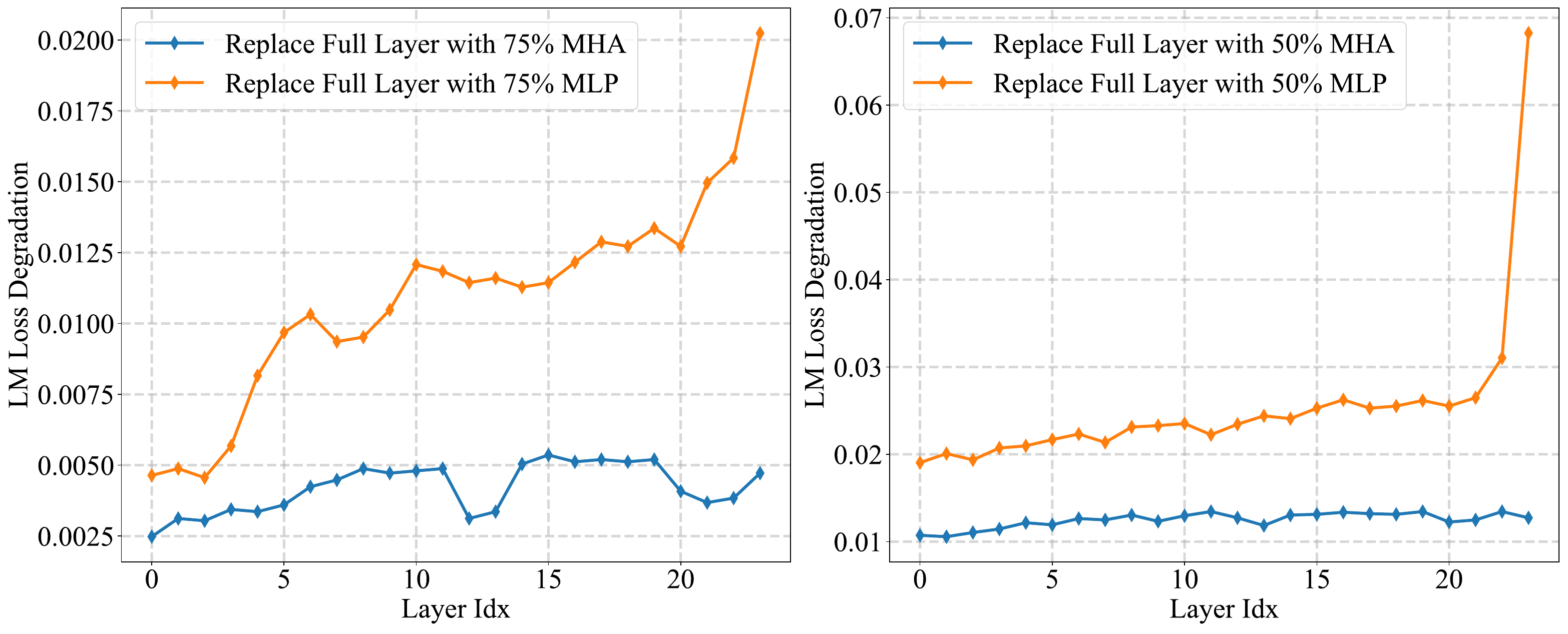}
    \vspace{-1em}
    \caption{Performance degradation introduced by replacing the full MHA/MLP layer with elastic candidates; specifically, the effect of replacing the full layer with $75\%$ and $50\%$ of the layer width. We observe that ($1$) using lightweight MHA layers could preserve more model performance, and ($2$) it's crucial to use full MLP layers deeper in the network. The experiment is based on GPT3-2B.}
    \vspace{-1.5em}
    \label{fig:vis_degradation}
\end{figure}

\begin{figure}[t]
    \centering
    \vspace{-1em}
    \begin{tabular}{c|c}
    50\% latency & 70\% latency \\
    \includegraphics[width=0.45\linewidth]{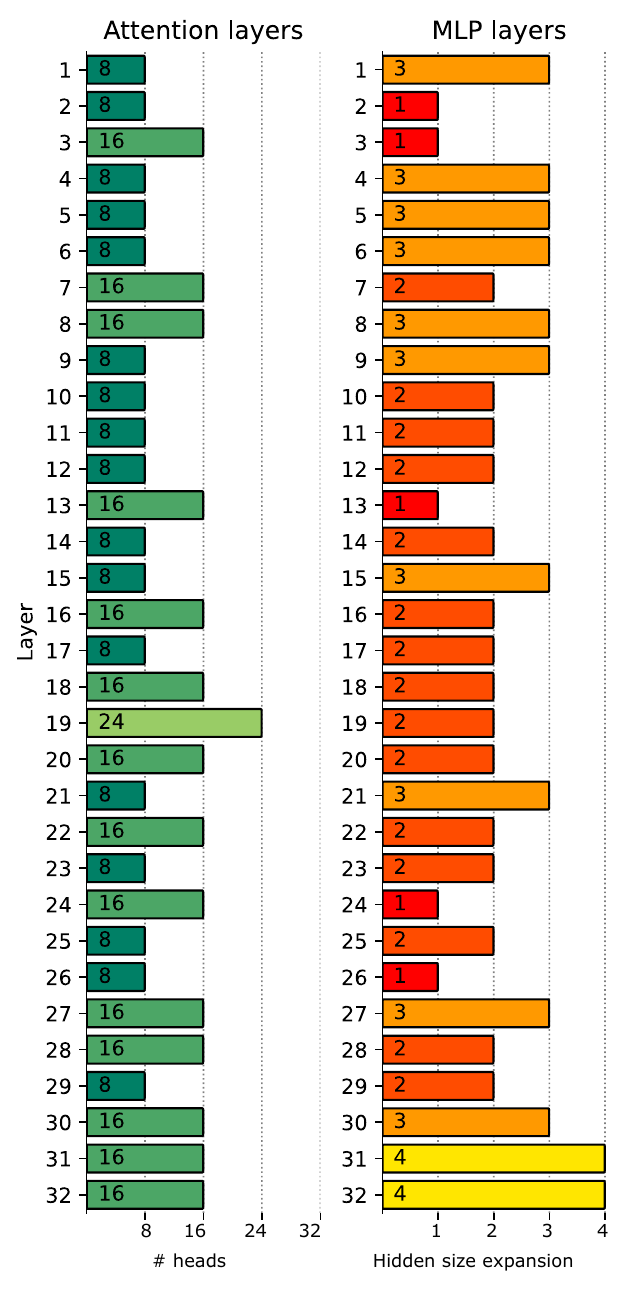}
    &
    \includegraphics[width=0.45\linewidth]{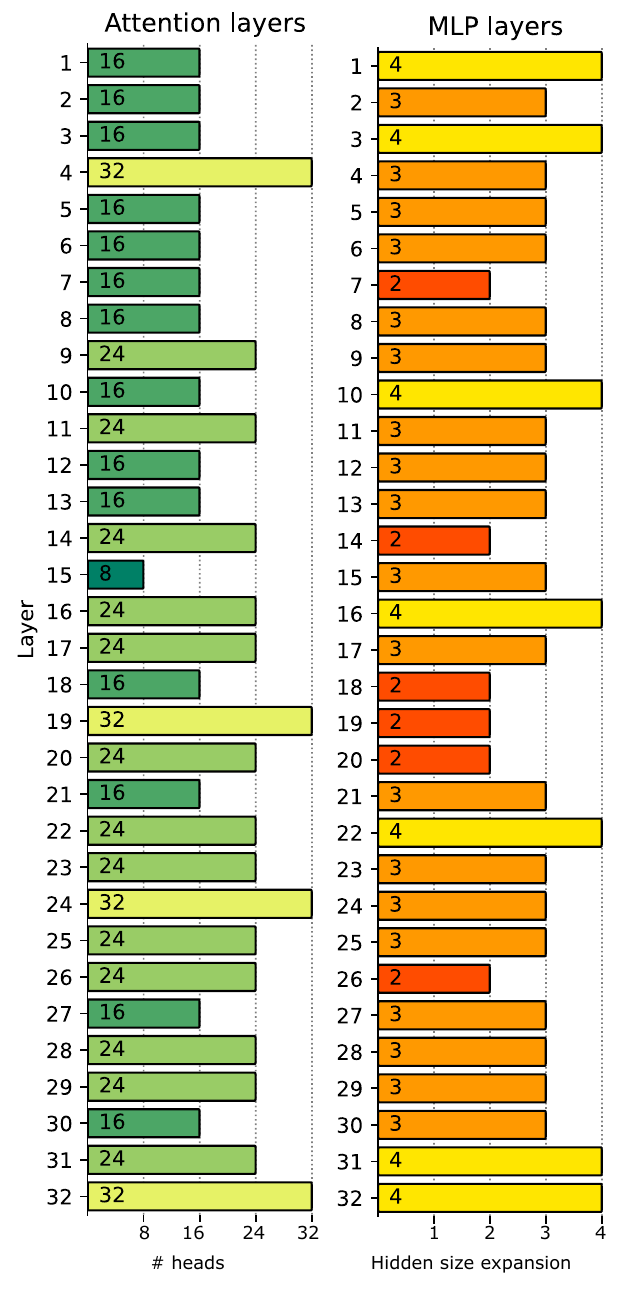}
    \end{tabular}
    \vspace{-1em}
    \caption{Obtained architectures for $50\%$ and $70\%$ latency targets.
    }
    \label{fig:expriments_archs}
    \vspace{-1.5em}
\end{figure}

During inference, MHA and MLP layers have similar latency, despite having different sizes in terms of parameters. For instance, in the GPT3-2B model, processing each MHA and MLP layer requires $3.830$ ms and $3.016$ ms, respectively.
In practical scenarios where low latency is crucial, understanding how to distribute compute and the number of model parameters among MLP and MHA layers becomes essential. \algoName provides us with a test-bed. In Figure~\ref{fig:vis_degradation}, for GPT3-2B model, we replace the full MHA/MLP layer with elastic candidates and calculate the performance degradation in terms of averaged LM loss. We compute the averaged LM loss over $7$ data domains, similar to previous sections. Two conclusions can be drawn: ($1$) replacing the full MLP layers results in higher performance degradation; ($2$) replacing deep layers, especially deep MLP layers, significantly hurts performance. 
Additionally, we visualize two Llama2-7B-based models with different latency targets, optimized by learnable static routers in Figure~\ref{fig:expriments_archs}. We observe that the learned structure aligns with the previous conclusions. We visualize other optimized architectures in the Appendix~\ref{sec:appendix:arch} and provide guidelines of architecture designs.

\paragraph{Input-adaptive Routers Assign More Computation to Hard Samples.}

\begin{figure}[htb]
    \centering
    \includegraphics[width=0.65\linewidth]{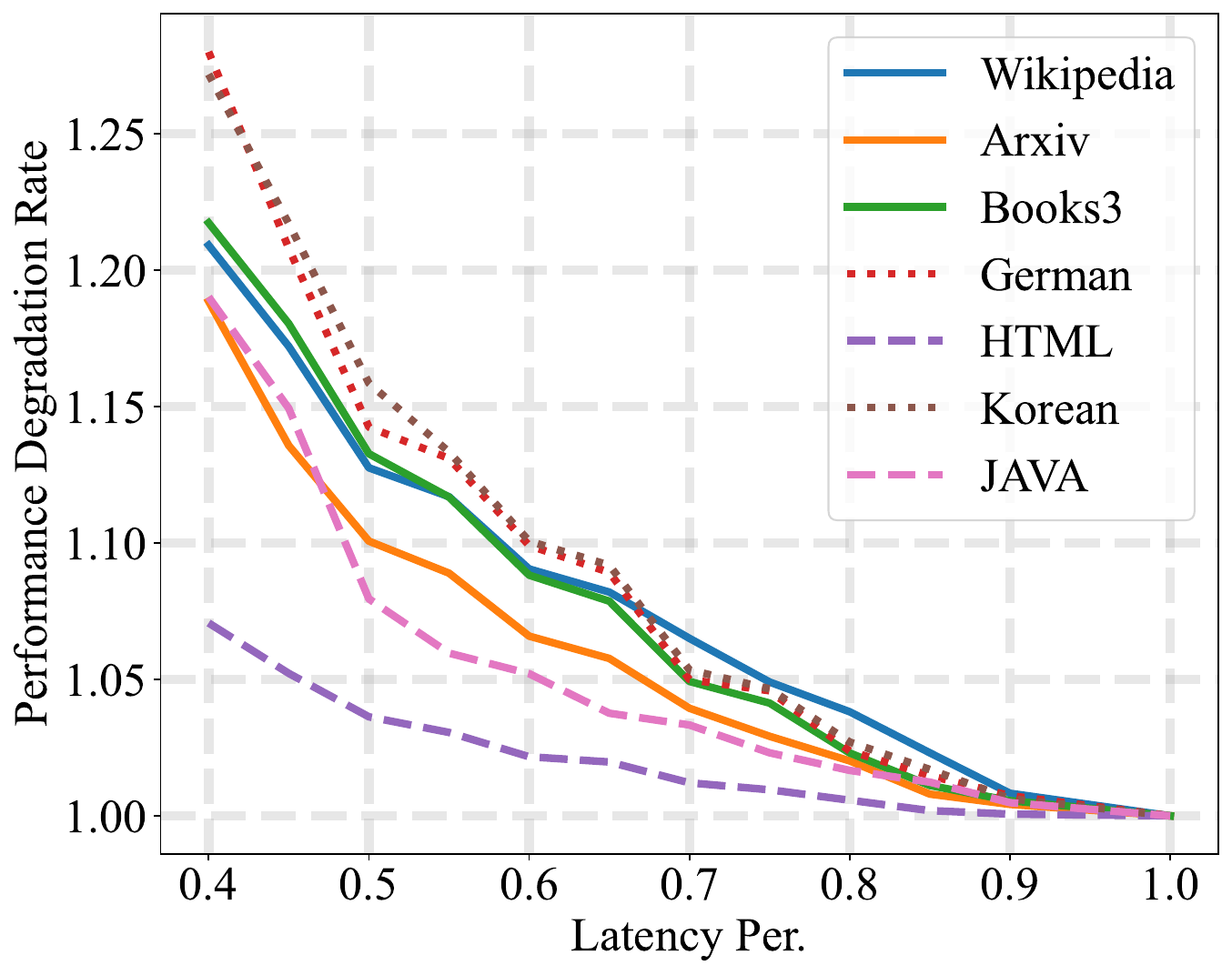}
    \vspace{-0.5em}
    \caption{Performance degradation
    on sub-networks of different latency, on different data domains.}
    \vspace{-0.5em}
    \label{fig:motivation}
\end{figure}

The necessity of input-adaptive routing naturally comes from data diversity. Typically, ``easy" datasets only need small-scale models for good performance, while ``hard" datasets require large-scale models. This observation motivated us to include support for input-adaptive routing in \algoName. We evaluate this hypothesis in Figure~\ref{fig:motivation}.
Here, we evaluate the sub-networks optimized by routers, with different latency, across multiple data domains. We mainly test on three categories: ($1$) English datasets: Arxiv, Books3~\cite{pile}, Wikipedia~\cite{wikidump}, ($2$) multilingual datasets: Korean, German, and ($3$) code data: HTML, JAVA. 
On GPT3-2B, we visualize the performance degradation of networks, calculated by $(\text{PPL}\textsubscript{sub}/\text{PPL}\textsubscript{full})$, and plot their correlation with latency. As a concrete example, notice that the curves for code datasets are much flatter than others, indicating that the task only requires a relatively small number of parameters. Conversely, multilingual datasets require more model parameters.

\begin{figure}[htb]
    \centering
    \includegraphics[width=0.9\linewidth]{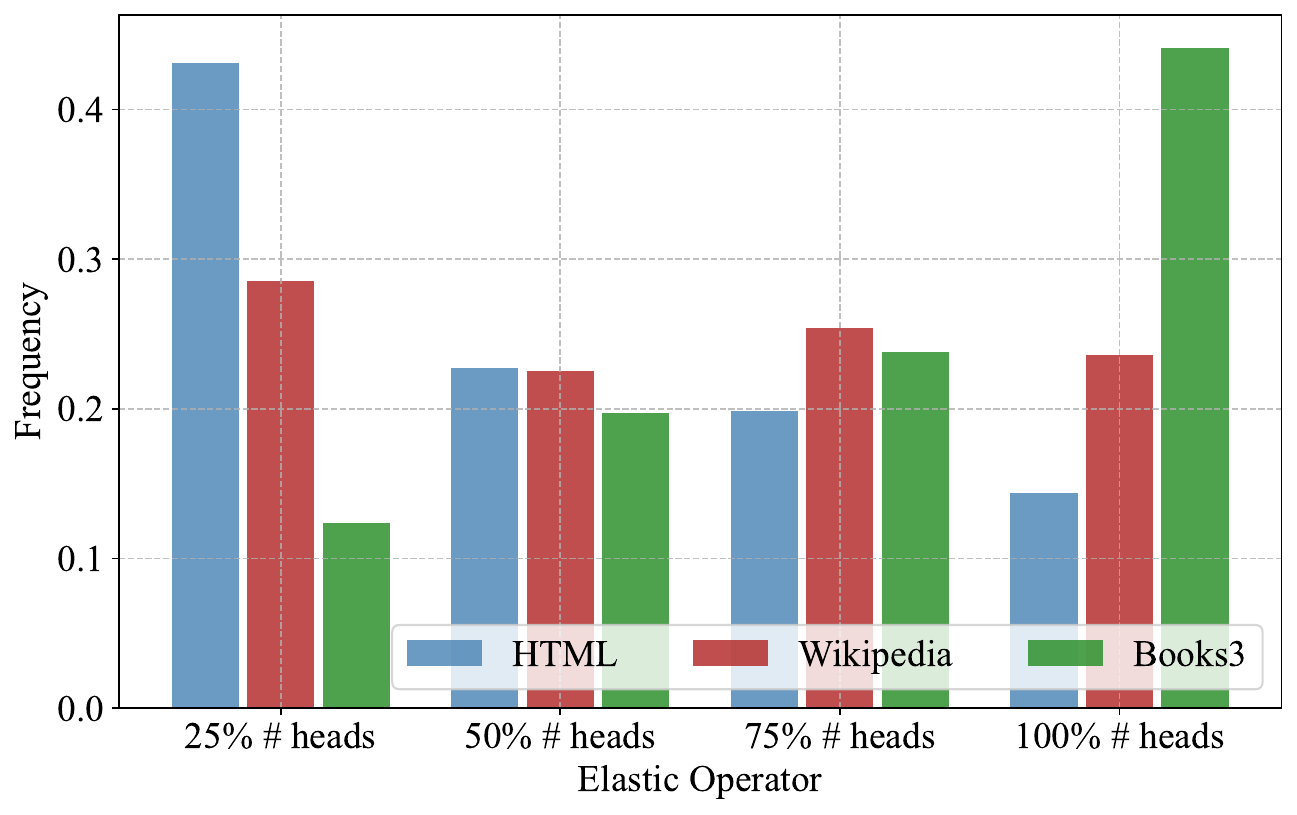}
    \caption{Router allocation vs data domain. 
    The y-axis depicts elastic operator remaining computations and horizontal axis depicts the frequency of the operator being chosen. 
    On GPT3-2B, we observe that ``hard data" (such as data from Books3~\cite{pile}, with a PPL of 11.64 on GPT3-2B) tend to utilize full layers more frequently. 
    Almost half of the tokens on ``easy data" (e.g., HTML dataset, with a PPL of 1.571) select the smaller layers. }
    \label{fig:histogram_dynamic}
\end{figure}

The input-adaptive models exhibit similar behavior. In Figure~\ref{fig:histogram_dynamic}, we selected the model with $61.8\%$ latency and obtained the router decision statistics for the first layer. For the HTML dataset, almost half of the tokens select the smallest elastic candidate, while tokens of the Books3 dataset~\cite{pile} tend to choose the full layer.

\subsection{Training Trajectory of Elastic Continued-Training}

\begin{figure}[htb]
    \centering
    \includegraphics[width=0.43\textwidth]{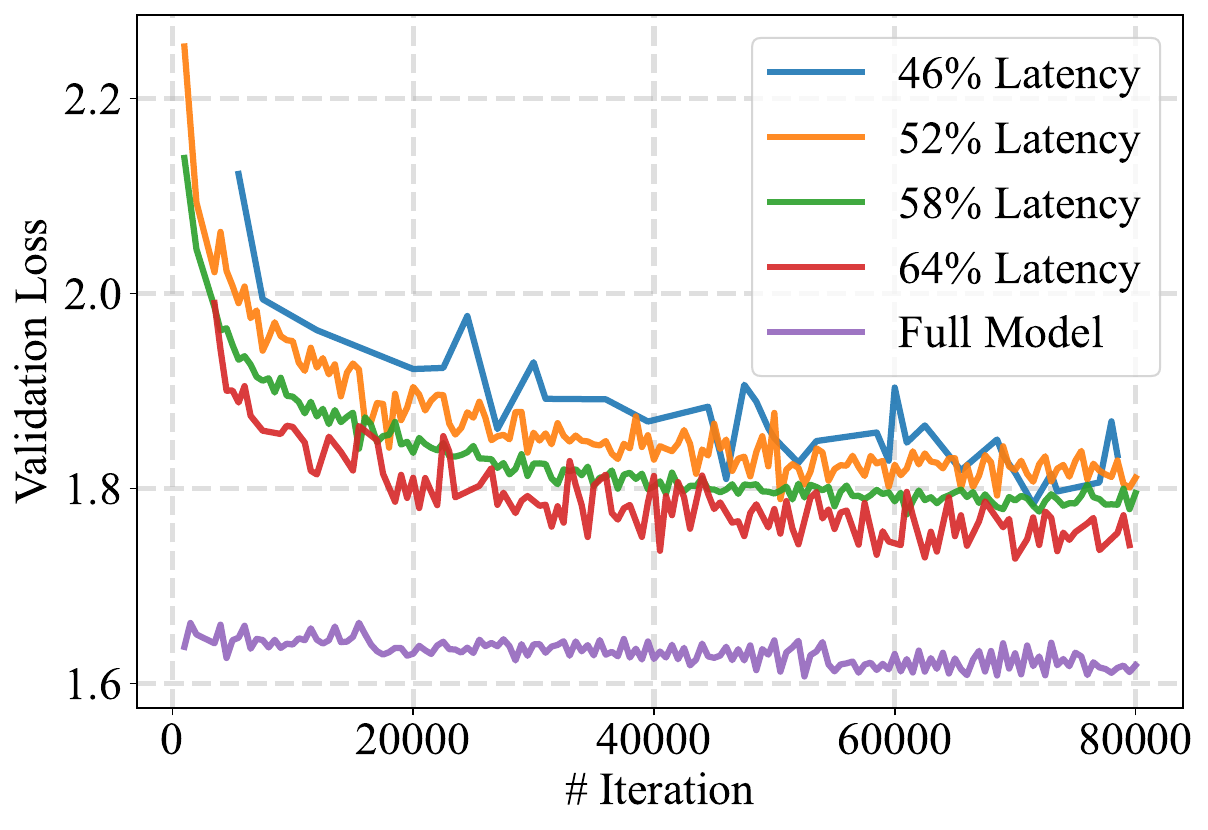}
    \vspace{-1em}
    \caption{Visualization of validation loss for sub-models of varying sizes during elastic continued-training.}
    \label{fig:training_trajectory_random}
\end{figure}

\definecolor{color1}{HTML}{1f77b4}
\definecolor{color2}{HTML}{ff7f0e}
\definecolor{color3}{HTML}{2ca02c}
\definecolor{color4}{HTML}{d62728}
\definecolor{color5}{HTML}{9467bd}
\definecolor{color6}{HTML}{8c564b}
In Figure~\ref{fig:training_trajectory_random}, we visualize the validation loss of sub-models of different sizes during the elastic continued-training process. We draw the following conclusions: 
($1$) elastic continued-training does not negatively impact the performance of the full model (i.e., employing all attention heads in MHA and full hidden size in MLP), as demonstrated by the \textcolor{color5}{purple} curve, ($2$) throughout training, all sub-networks converge synchronously, while the larger sub-models lead to smaller validation losses overall. To validate, we depict validation loss of randomly selected sub-networks using the \textcolor{color1}{blue}, \textcolor{color2}{orange}, \textcolor{color3}{green}, and \textcolor{color4}{red} curves, incurring $46\%$, $52\%$, $58\%$, $64\%$ of the full latency, respectively. Note that the sub-models are randomly picked independently at each validation step, ($3$) the middle-sized sub-models exhibit more stable convergence, as indicated by the smoother curves for the \textcolor{color3}{green} and \textcolor{color2}{orange} lines, compared to the \textcolor{color1}{blue} and \textcolor{color4}{red} ones. This stability could potentially be attributed to the fact that middle-sized models are sampled more frequently during elastic continued-training.

\begin{figure*}[htb]
    \centering
    \includegraphics[width=\linewidth]{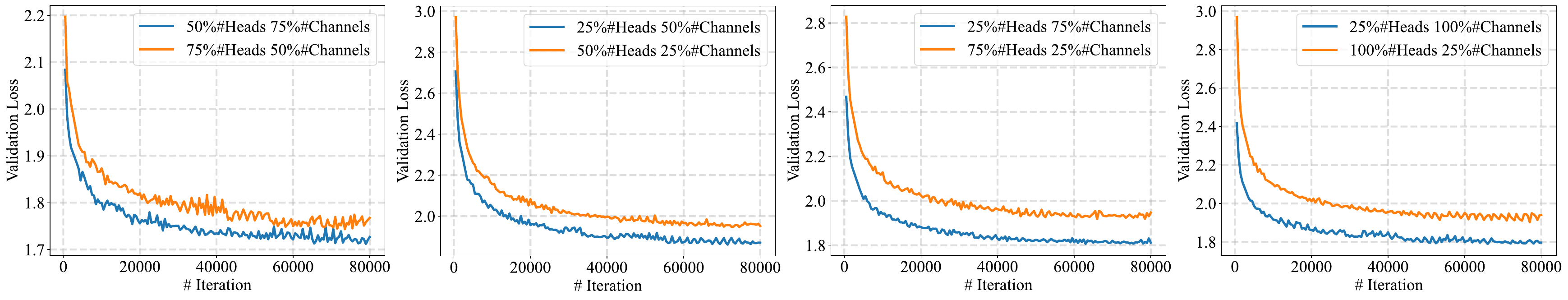}
    \vspace{-2em}
    \caption{Training trajectory of models performing uniform elastic selection strategy.}
    \label{fig:experiment:uniform_curve}
\end{figure*}

We also show the trajectory of the validation loss of the models employing a uniform elastic selection strategy in Figure~\ref{fig:experiment:uniform_curve}. For instance, in the first sub-figure, the ``$50\%\#$Heads $75\%\#$Channels'' refers to the model selecting the first half of the attention heads and $75\%$ of the channels for all layers. The figure echoes the observation in Section~\ref{sec:insights} that the adoption of lightweight MHAs, characterized by a reduced number of heads, is more advantageous in limited-resource regimes.

\subsection{Router Training Dynamics.}

\begin{figure}[htb]
    \centering
    \vspace{-1em}
    \includegraphics[width=0.43\textwidth]{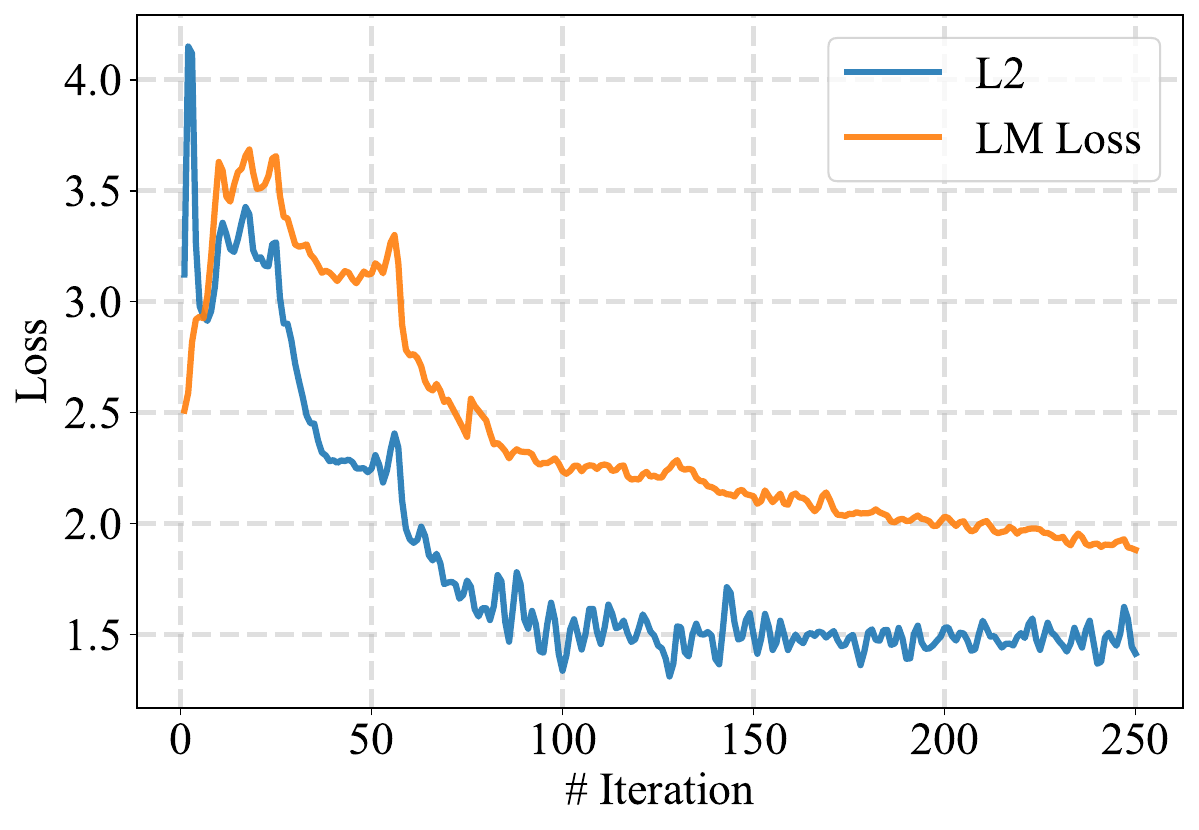}
    \vspace{-1em}
    \caption{Router training dynamics. We visualize the curve of losses (``L2 Loss'' and ``LM Loss'') during the router training. }
    \label{fig:experiment:router_dynamics}
\end{figure}

During router training, we introduce the surrogate model (SM), to estimate the language modeling loss (LM Loss) based on router logits, providing a stable signal for router training. As detailed in Figure~\ref{fig:train_router}, when the SM is not accurate enough, the ``L2 Loss'' is utilized, where ``L2 Loss'' refers to the MSE loss between the ``ground truth'' language modeling loss and the estimated LM loss via SM. When the SM error is smaller than the threshold, the router will be optimized based on the estimated LM loss.
We depict the dynamics of router training in Figure~\ref{fig:experiment:router_dynamics}. Router training can be roughly divided into three stages: ($1$) SM tuning: the ``L2 Loss'' quickly drops, during which the LM Loss slightly increases; ($2$) Joint tuning: both losses decrease simultaneously; ($3$) Router tuning: the LM Loss continues to decrease while the "L2 Loss" remains below the threshold.

\subsection{Effectiveness of Learned Routers}
\begin{figure}[htb]
    \centering
    \includegraphics[width=0.7\linewidth]{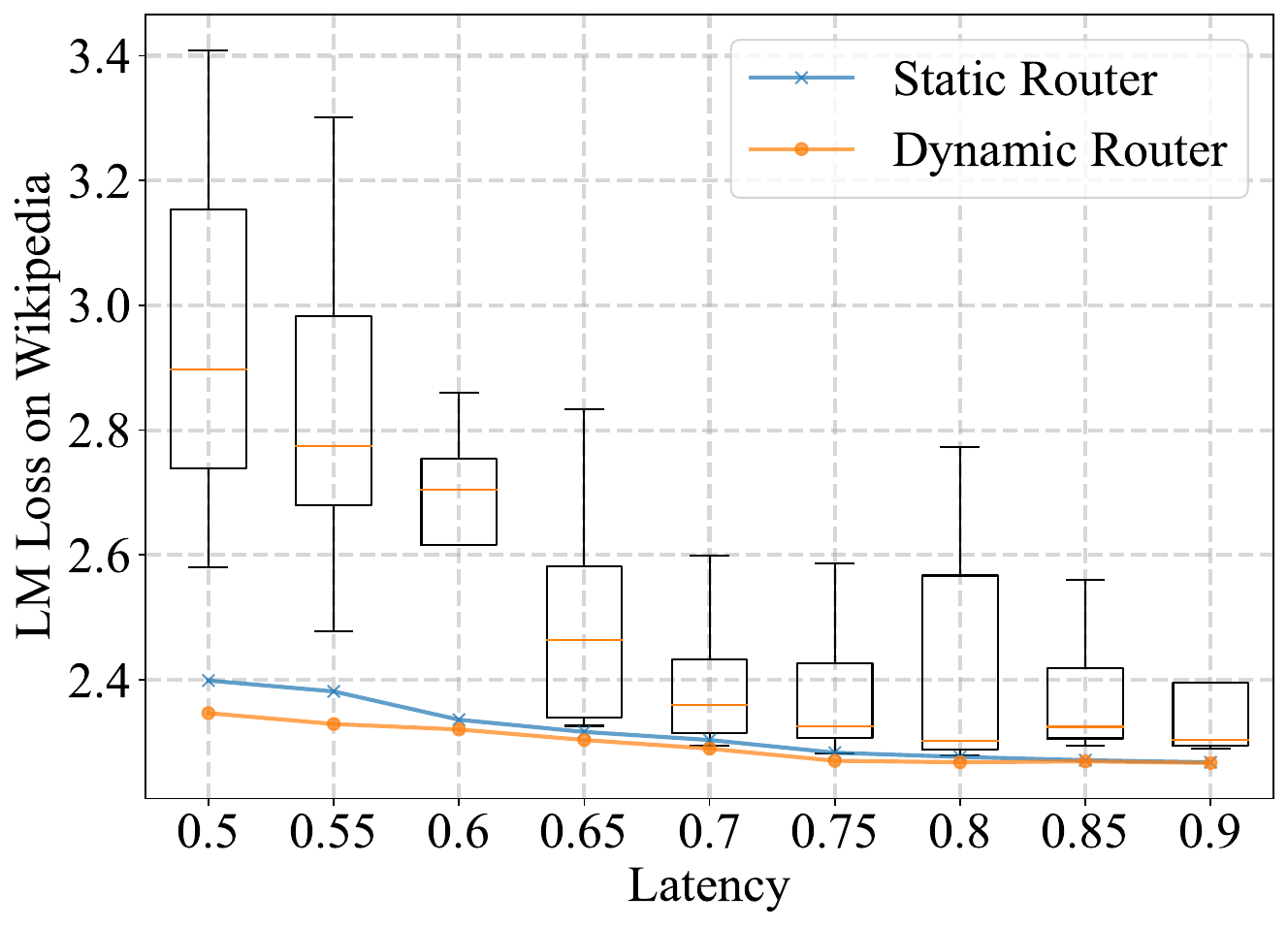}
    \vspace{-0.5em}
    \caption{Effectiveness of our automatic network selection algorithm. The box-plot visualizes the performance distribution of randomly selected models. The blue and yellow lines denote performance of \algoName's routers. Performance is evaluated on Wikipedia~\cite{wikidump} and GPT3-2B Flextron models.}
    \label{fig:bloxplot}
\end{figure}

To demonstrate the effectiveness of our learned routers, we compare the learned sub-models with randomly picked ones. In Figure~\ref{fig:bloxplot}, we first randomly sample sub-models at different latencies from GPT3-2B, and measure their performance. We use box plots to visualize the distributions of their LM loss. As seen from the Figure, the majority of randomly selected sub-models have unpredictable performance. For instance, sub-models at $65\%$ latency have averaged LM loss ranging from $2.32$ to $3.06$. We compare them to sub-models found by routers (blue and yellow lines in the Figure), demonstrating that \algoName effectively identifies optimal sub-models.

\subsection{Effectiveness and Necessity of Weight Permutation}\label{sec:analysis:ranking}

We assess the effectiveness of weight permutation by testing the performance of the original and permuted models using their first-half (50\%) MLP neurons/MHA heads as is. As Table~\ref{table:zero_permute} demonstrates, the perplexity on Wikipedia~\cite{wikidump} significantly improves post-permutation, with the un-permuted model's perplexity exceeding $1000$, while the permuted model's perplexity was $193.6$. We observe a similar enhancement for MHA modules.

\begin{table}[htb]
\caption{Ablation on permuting the base model by channel/neuron importance score (Eq.~\ref{eq:permute_head} and Eq.~\ref{eq:permute_mlp} in Section~\ref{sec:method:training}) as the initialization. Numbers correspond to the Wikipedia perplexity of pre-trained models cut to the first half neurons/heads. Note that we report the zero-shot performance of the permuted model.
\label{table:zero_permute}}
\centering
\resizebox{.9\columnwidth}{!}{
\begin{tabular}{@{}lrr@{}}
\toprule
Setup & MHA & MLP \\ \midrule
Full (baseline) & $9.144$ & $9.144$ \\ \midrule
$50\%$ Operator w/o Elastic Sorting & $184.5$ & $1902.0$ \\
$50\%$ Operator w/ Elastic Sorting (ours) & $179.9$ & $193.6$ \\
\bottomrule
\end{tabular}
}
\end{table}

\section{Related Work}

\paragraph{Elastic Inference.}
 
The idea of obtaining multiple models from a single trained model has  been explored extensively in the convolutional neural network (CNN) literature;
in particular, \citet{yu2018slimmable, yu2019universally} introduced slimmable neural networks, which support deployment of the same model with varying numbers of convolutional filters. \citet{li2021dynamic} leverage a gating mechanism to dynamically identify sample difficulty and adjust the percentage of activated filters accordingly. Finally,~\citet{cai2019once} generalized pruning methods to derive a single model adaptable to different configurations. 
Recent work has explored the application of slimmable models to Transformer-derived architectures; specifically,~\citet{rao2021dynamicvit} and~\citet{yin2022vit} explore the mechanism of slimmable token removal for adaptive token dropping. ~\citet{kusupati2022matryoshka} introduce a nested weight structure for Transformer networks, and~\citet{kudugunta2023matformer} use this formalization in the Matformer architecture.
\citet{valipour2023sortednet} additionally utilize a sampling-based training strategy to train multiple models via gradient accumulation.
While \algoName shares Matformer's nested weight structure, it uniquely extends it by offering elasticity in both MLP and MHA layers, a larger pool of operations, efficient pretraining for sub-linear training times, and automatic input-adaptive sub-network selection based on latency for enhanced efficiency.

\paragraph{Input Adaptivity.}
Sparse Mixture-of-Expert networks (MoEs) utilize input adaptivity to achieve efficient model scaling by collectively utilizing multiple specialized sub-networks~\cite{fedus2022switch, riquelme2021scaling, zhou2022mixture, jiang2024mixtral}, to handle data from diverse domains~\cite{li2022branch,zhang2023robust}. Tokens in MoE networks only pass through the most relevant sub-networks, identified by learnable routers. 
Recent work has started challenging the traditional definition of MoEs by introducing heterogeneous experts~\cite{wang2020hat,dean2021introducing,zhou2022mixture} and in-situ adaptiveness~\cite{chen2023sparse,cai2023robust}.
However, all existing MoE designs that we are aware of store expert weights separately, with no notion of weight sharing.
This design introduces significant memory and communication overheads, especially at larger batch sizes with higher expert utilization. In \algoName, all the ``experts'' in a layer share the same weight matrix, and different sub-networks are selected through simple array indexing, thus relieving most of the pressure from the memory and networking interconnect.
Additionally, \algoName includes provisions for routing decisions to be dictated by a latency target, a feature absent from most existing MoE networks.

\paragraph{Static Acceleration.} A vast body of work has also demonstrated the efficacy of static acceleration methods on transformers, including weight and activation quantization~\citet{lin2023awq, frantar2022gptq}, patterned 2:4 sparsity~\citet{mishra2021accelerating}, neural architecture search (NAS)~\citet{wang2020hat,wu2021autoformer}, and hardware-aware structural pruning~\citet{yang2023global}. Besides, \citet{ma2023llm,xia2023sheared,wang2023learning,wang2023data} aim to re-use pre-trained checkpoints to avoid repeated computation. These methods are orthogonal to the dynamic inference literature and can provide further opportunities for performance improvement.

\section{Conclusion}
This paper has presented \algoName, a novel network architecture and post-training optimization framework. \algoName models flexibly adapt to different latency and accuracy targets during inference with no additional fine-tuning, and come with built-in support for input-adaptive routing to maximize performance. We have also presented a post-training framework for systematically converting standard trained LLMs such as GPT-3 and Llama2 into \algoName models using a sample-efficient training procedure.
\algoName demonstrates superior zero-shot performance over multiple smaller end-to-end trained variants on the GPT-3 family and Llama-2-7B model; \algoName also outperforms the state-of-the-art Matformer framework~\cite{kudugunta2023matformer}. \algoName achieves this through a single pretraining run that consumes a mere $7.63\%$ of training tokens of full pretraining cost.

\bibliography{paper}
\bibliographystyle{icml2024}

%%%%%%%%%%%%%%%%%%%%%%%%%%%%%%%%%%%%%%%%%%%%%%%%%%%%%%%%%%%%%%%%%%%%%%%%%%%%%%%
%%%%%%%%%%%%%%%%%%%%%%%%%%%%%%%%%%%%%%%%%%%%%%%%%%%%%%%%%%%%%%%%%%%%%%%%%%%%%%%
% APPENDIX
%%%%%%%%%%%%%%%%%%%%%%%%%%%%%%%%%%%%%%%%%%%%%%%%%%%%%%%%%%%%%%%%%%%%%%%%%%%%%%%
%%%%%%%%%%%%%%%%%%%%%%%%%%%%%%%%%%%%%%%%%%%%%%%%%%%%%%%%%%%%%%%%%%%%%%%%%%%%%%%

\newpage
\appendix
\onecolumn

\section{Architecture Visualization}\label{sec:appendix:arch}

\begin{wrapfigure}{R}{\textwidth}
    \centering
    \begin{tabular}{c|c|c|c}
    \multicolumn{2}{c}{GPT3-2B} & \multicolumn{2}{c}{GPT3-8B} \\
    50\% latency & 70\% latency & 50\% latency & 70\% latency \\
    \includegraphics[height=0.45\linewidth]{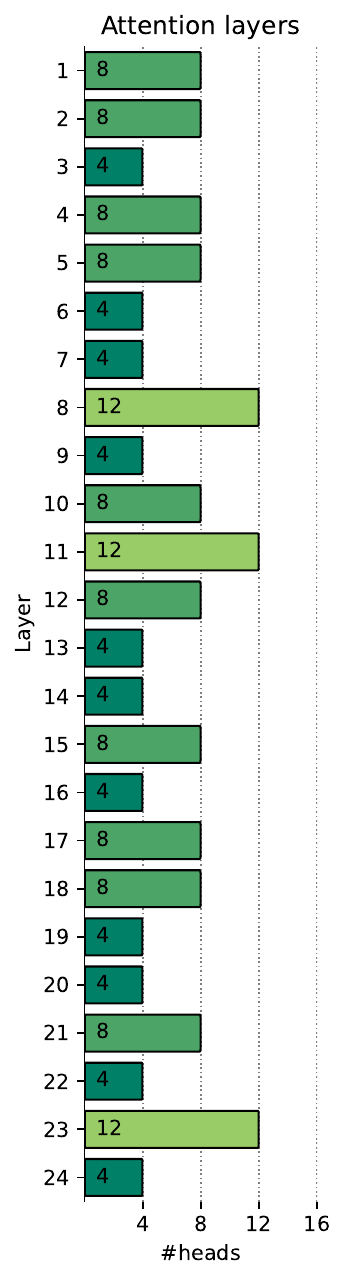}
    \includegraphics[height=0.45\linewidth]{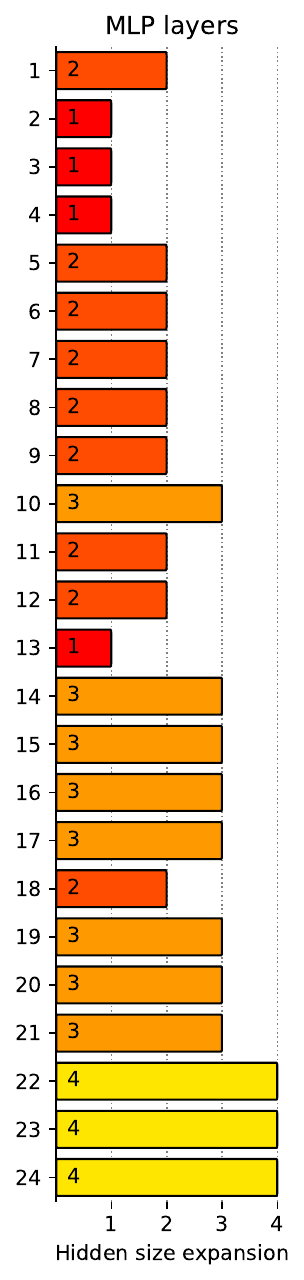}
    &
    \includegraphics[height=0.45\linewidth]{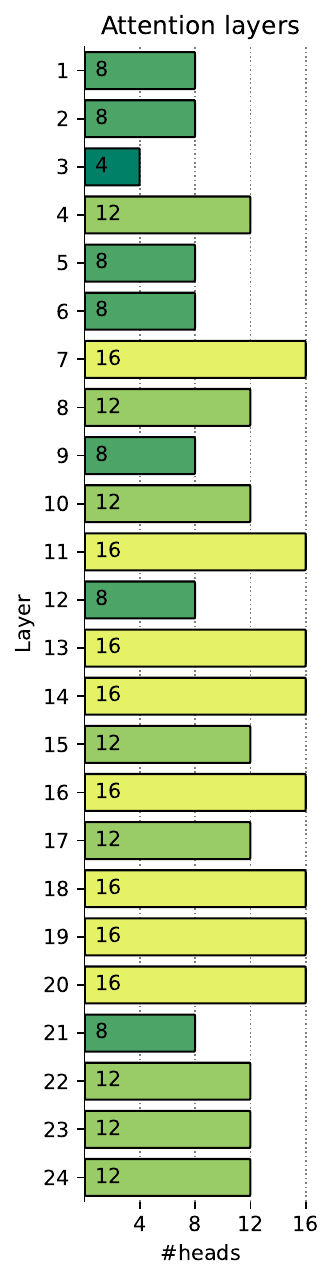}
    \includegraphics[height=0.45\linewidth]{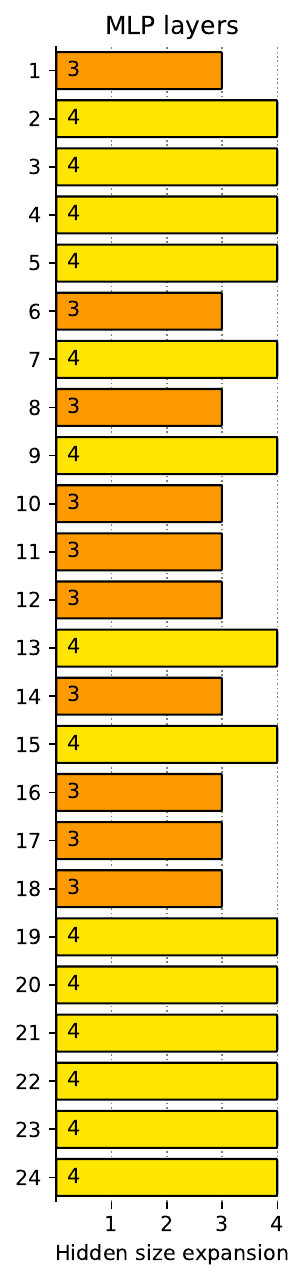}  
    &
    \includegraphics[height=0.45\linewidth]{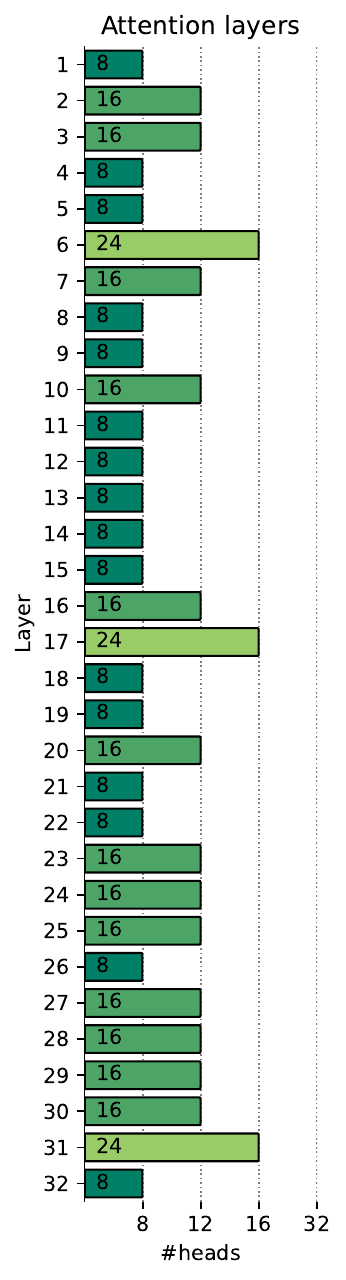}
    \includegraphics[height=0.45\linewidth]{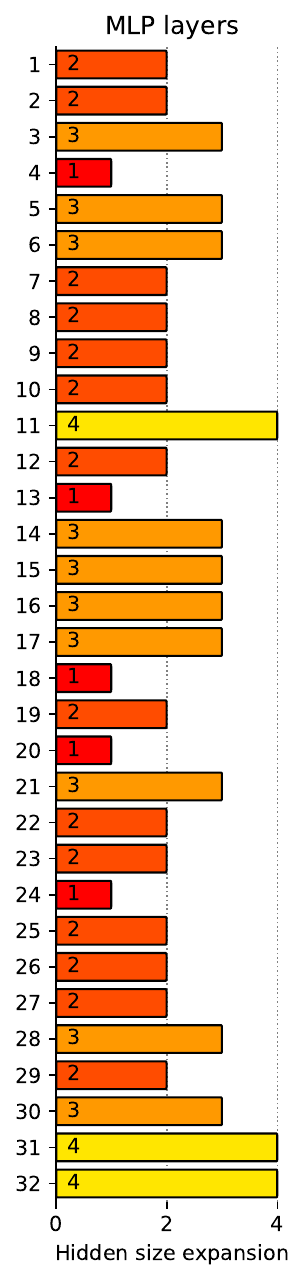}
    &
    \includegraphics[height=0.45\linewidth]{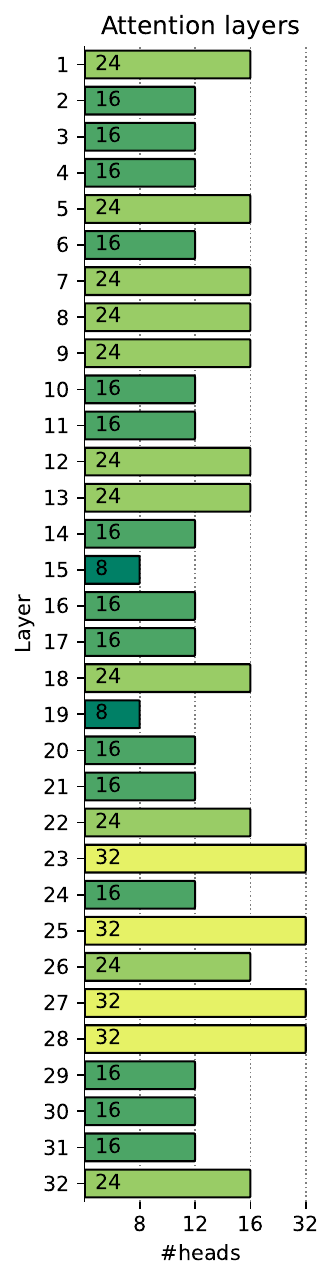}
    \includegraphics[height=0.45\linewidth]{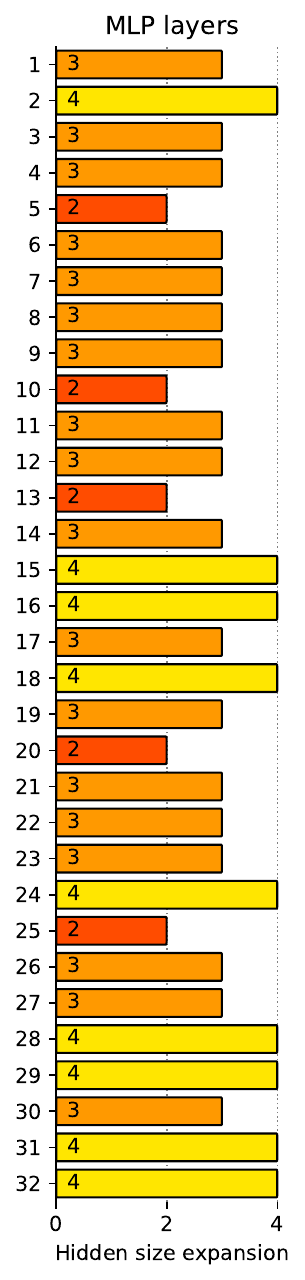}
    \end{tabular}
    \caption{Obtained architectures for $50\%$ and $70\%$ latency targets based on GPT-3 family.}
    \label{fig:expriments_archs_app}
\end{wrapfigure}

We provide searched architectures in Figure~\ref{fig:expriments_archs_app}, based on two variants of the GPT3 family. The observation validates our previous heuristic entailed in Sec~\ref{sec:insights}.

%You can have as much text here as you want. The main body must be at most $8$ pages long. For the final version, one more page can be added. If you want, you can use an appendix like this one.  

%The $\mathtt{\backslash onecolumn}$ command above can be kept in place if you prefer a one-column appendix, or can be removed if you prefer a two-column appendix.  Apart from this possible change, the style (font size, spacing, margins, page numbering, etc.) should be kept the same as the main body.
%%%%%%%%%%%%%%%%%%%%%%%%%%%%%%%%%%%%%%%%%%%%%%%%%%%%%%%%%%%%%%%%%%%%%%%%%%%%%%%
%%%%%%%%%%%%%%%%%%%%%%%%%%%%%%%%%%%%%%%%%%%%%%%%%%%%%%%%%%%%%%%%%%%%%%%%%%%%%%%

\end{document}